\documentclass{CUP-JNL-DTM}%

\usepackage{graphicx}
\usepackage{multicol,multirow}
\usepackage{amsmath,amssymb,amsfonts}
\usepackage{mathrsfs}
\usepackage{amsthm}
\usepackage{rotating}
\usepackage{appendix}
\usepackage[numbers]{natbib}
\usepackage{ifpdf}
\usepackage[T1]{fontenc}
\usepackage[utf8]{inputenc}

\usepackage{newtxtext}
\usepackage{newtxmath}
\usepackage{textcomp}
\usepackage[table, dvipsnames]{xcolor}
\usepackage{lipsum}
\usepackage{changepage}
\usepackage[pagebackref=true,breaklinks=true,colorlinks,citecolor=blue,bookmarks=false]{hyperref}
\usepackage{arydshln}

\definecolor{lightgray}{gray}{0.9}

\theoremstyle{definition}

\numberwithin{equation}{section}

\jname{Data/Math}
\articletype{ARTICLE TYPE}
\jyear{2023}

\begin{document}
\UseRawInputEncoding
\begin{Frontmatter}
\title[Article Title]{OpenForest: A data catalogue for machine learning in forest monitoring}

\author[1,2]{Arthur Ouaknine}
\author[3,4]{Teja Kattenborn}
\author[5]{Etienne Lalibert\'e}
\author[1,2]{David Rolnick}

\authormark{Ouaknine \textit{et al}.}

\address[1]{\orgdiv{School of Computer Science}, \orgname{McGill University}, \orgaddress{\city{Montr\'eal}, \postcode{H3A 2A7}, \state{Qu\'ebec},  \country{Canada}}}

\address[2]{\orgdiv{Mila}, \orgname{Quebec AI Institute}, \orgaddress{\city{Montr\'eal}, \postcode{H2S 3H1}, \state{Qu\'ebec},  \country{Canada}}. \email{arthur.ouaknine@mila.quebec}}

\address[3]{\orgdiv{Remote Sensing Centre for Earth System Research}, \orgname{Leipzig University}, \orgaddress{\city{Leipzig}, \postcode{04109}, \country{Germany}}}

\address[4]{\orgdiv{German Centre for Integrative Biodiversity Research (iDiv)}, \orgaddress{\city{Halle-Jena-Leipzig}, \postcode{04103}, \country{Germany}}}

\address[5]{\orgdiv{Institut de recherche en biologie v\'eg\'etale, D\'epartement de sciences biologiques}, \orgname{Universit\'e de Montr\'eal}, \orgaddress{\city{Montr\'eal}, \postcode{H1X 2B2}, \state{Qu\'ebec},  \country{Canada}}}



\abstract{
Forests play a crucial role in Earth's system processes and provide a suite of social and economic ecosystem services, but are significantly impacted by human activities, leading to a pronounced disruption of the equilibrium within ecosystems.
Advancing forest monitoring worldwide offers advantages in mitigating human impacts and enhancing our comprehension of forest composition, alongside the effects of climate change.
While statistical modeling has traditionally found applications in forest biology, recent strides in machine learning and computer vision have reached important milestones using remote sensing data, such as tree species identification, tree crown segmentation and forest biomass assessments.
For this, the significance of open access data remains essential in enhancing such data-driven algorithms and methodologies.
Here, we provide a comprehensive and extensive overview of 86 open access forest datasets across spatial scales, encompassing inventories, ground-based, aerial-based, satellite-based recordings, and country or world maps.
These datasets are grouped in \textbf{OpenForest}, a dynamic catalogue open to contributions that strives to reference all available open access forest datasets.
Moreover, in the context of these datasets, we aim to inspire research in machine learning applied to forest biology by establishing connections between contemporary topics, perspectives and challenges inherent in both domains.
We hope to encourage collaborations among scientists, fostering the sharing and exploration of diverse datasets through the application of machine learning methods for large-scale forest monitoring.
\textbf{OpenForest} is available at the following url: \url{https://github.com/RolnickLab/OpenForest}.
}

\end{Frontmatter}

\section*{Impact Statement}

OpenForest establishes a constantly evolving catalogue of open access forest datasets. This catalogue is open for contributions and aims to provide a single central hub for such datasets within the open-source community.
In addition to introducing the OpenForest catalogue, we provide in this paper a detailed overview of complementary research topics and challenges in forest monitoring and machine learning, so as to better enable the impactful use of these datasets in interdisciplinary research.
We hope this work will ultimately contribute substantially to enhancing our comprehension of global forest composition as well the development of innovative machine learning algorithms.


\localtableofcontents

\section[Introduction]{Introduction}

\begin{figure}[t]%
\FIG{\includegraphics[width=1\textwidth]{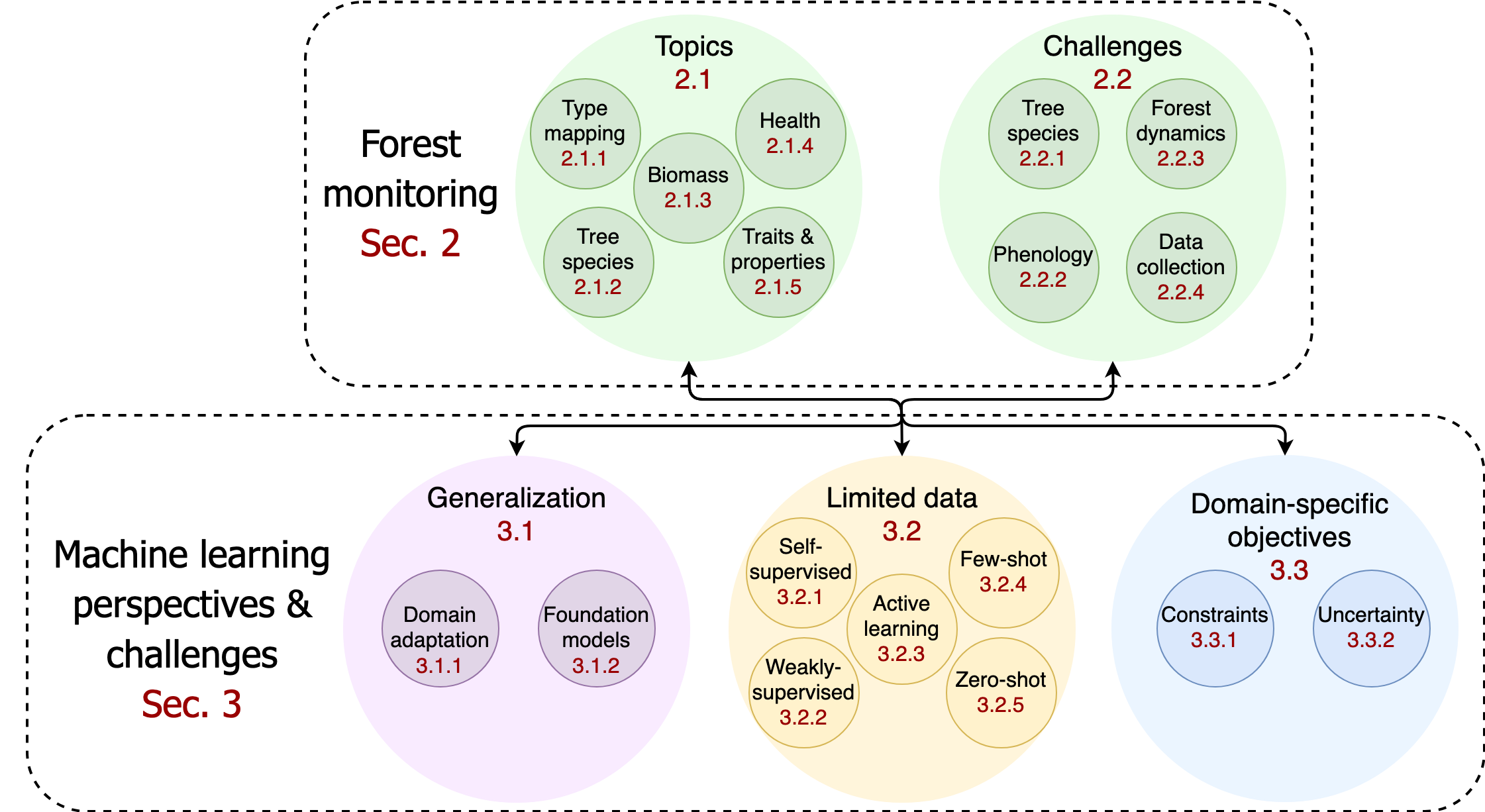}}
{\caption{\textbf{Overview of forest monitoring topics and challenges associated to machine learning perspectives and challenges.} Each forest monitoring topics and challenges are detailed with their corresponding section number (in red). They are associated to the three main machine learning perspectives and challenges categories, namely generalization, limited data and domain-specific objectives, alongside with their corresponding section number (in red)}
\label{fig:challenges}}
\end{figure}

Forests cover one third of the Earth's land surface \cite{the_food_and_agriculture_organization_of_the_united_nations_global_2020}.
They provide a range of valuable ecosystem goods and services to humanity, including timber provision, water and climate regulation, and atmospheric carbon sequestration \cite{assessment_millennium_2001, bonan_forests_2008}. 
They also serve as habitat for a myriad of plant, animal, and microbial species. 
However, human activities have had, and continue to have, a major impact on forests worldwide. 

More than 3000 ha of forests disappear every hour from deforestation \cite{the_food_and_agriculture_organization_of_the_united_nations_global_2020, hansen_high-resolution_2013}.
Yet forests are also increasingly recognized as natural solutions to the joint climate and biodiversity crises \cite{griscom_natural_2017, griscom_national_2020, drever_natural_2021}.
Forest-based adaptation through avoided forest conversion, improved forest management, and forest restoration could mitigate over 2 Gt CO\textsubscript{2}-eq emissions per year by 2030 \cite{intergovernmental_panel_on_climate_change_ipcc_climate_2023},
with variations observed in different regions worldwide \cite{griscom_natural_2017, bastin_global_2019, busch_potential_2019, griscom_national_2020, drever_natural_2021}, while being limited by the climatic effects we are witnessing on forests \cite{Zhu2018}.

Due to their significant economic and ecological importance, monitoring forests has attracted considerable attention. 
It includes the assessment of ecosystem functional properties, as well as the evaluation of forest health, vitality, stress, and diseases (see Sec.~\ref{sec:forest_topics}).
However, monitoring forests presents significant challenges, especially using field-based approaches (see Sec.~\ref{sec:forest_challenges}). 
Forests cover huge areas, and can be difficult to access. 
Consequently, remote sensing has and continues to play an important role in forest monitoring worldwide. 
Nowadays, a wide array of remote sensing platforms and sensors to monitor forests are available and being used. 
This includes platforms such as drones (also referred to as unoccupied aerial vehicles or UAVs), airplanes, or satellites, and sensors ranging from passive optical imagery, to active methods such as light detection and ranging (LiDAR) or synthetic aperture RADAR (SAR) \citep{white_remote_2016, verrelst_optical_2015}.

In recent years, the applications of remote sensing data for Earth-related purposes \cite{campsvalls_deep_2021, ma_deep_2019}, such as forest monitoring \cite{fassnacht_review_2016, kattenborn_review_2021, diez_deep_2021, michalowska_review_2021}, have gained momentum due to the adoption of machine learning methods and algorithms. 
It has been inspired by continuous improvements in the performance of deep learning models used in computer vision challenges in the past decade \cite{deng_imagenet_2009,everingham_pascal_2015,lin_microsoft_2014}.
Recent advances in deep learning model architectures have enabled the integration of remote sensing data from various sensors and resolutions - spatial, temporal, or spectral - which presents promising opportunities to enhance forest monitoring practices \cite{rahaman_general_2022, cong_satmae_2022, reed_scale-mae_2022, tseng_lightweight_2023}.

Numerous machine learning challenges related to forest monitoring have yet to be explored (see Fig.~\ref{fig:challenges}), and addressing them will require diverse and large forest datasets \cite{liang_importance_2020}. 
While there is a wealth of remote sensing data that is freely available, these data can be difficult to access because they involve a wide variety of sensory modalities, geographies, and tasks and are spread out across many repositories. To our knowledge, no comprehensive, central repository of open access forest datasets currently exists, a gap which we fill here with \textbf{OpenForest}\footnote{\url{https://github.com/RolnickLab/OpenForest}}. The \textbf{OpenForest} catalogue is designed to simplify the process of accessing and highlight forest monitoring datasets for researchers in the field of machine learning and forest biology, thereby accelerating progress in these domains.

\begin{figure}[t]%
\includegraphics[width=1.0\textwidth]{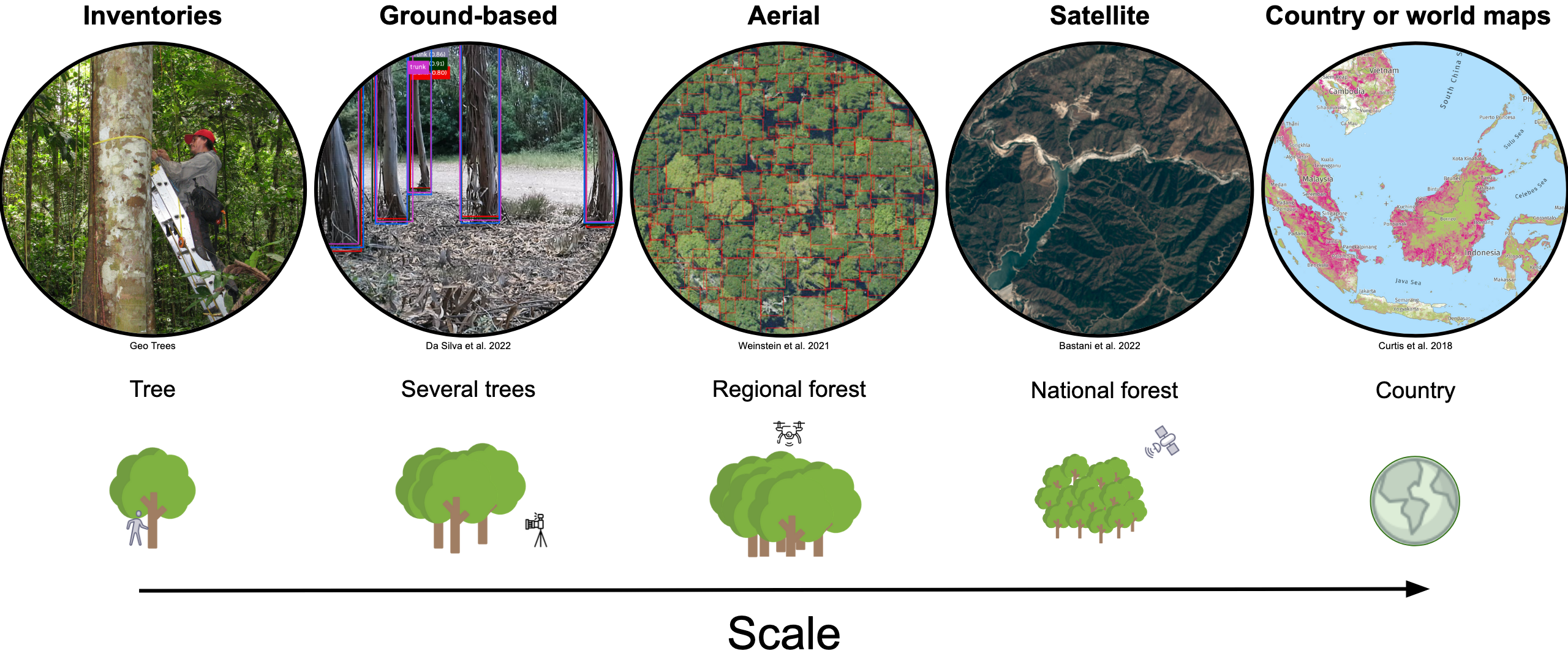}
\caption{
\textbf{Illustration of forest monitoring datasets at different scales.} Inventories are in situ measurements realised at the tree level. Ground-based datasets are recorded within or below the canopy of the trees. Aerial datasets are composed of recordings from sensors mounted on unoccupied (drones) or occupied aircrafts. Satellite datasets are collected from sensors mounted on satellites orbiting the Earth. Map datasets are generated at the country or world level using datasets at the aerial or satellite scales}
\label{fig:scale}
\end{figure}

In this paper, we present the existing biological topics and challenges related to forest monitoring that scientists are currently investigating (Sec.~\ref{sec:forest_topics_and_challenges}) and which could be of interest of machine learning practitioners. 
Additionally, we briefly introduce several machine learning research topics, exploring their potential applications in addressing biology-related challenges (Sec. \ref{sec:ml_challenges}). These applications hold promise in assisting ecologists and biologists in their work.
Moreover, we conduct a thorough review of open access forest datasets across different spatial scales (Fig.~\ref{fig:scale}) to support both machine learning and biology research communities in their work (Sec.~\ref{sec:review}). 
Finally, we provide perspectives on the space of machine learning applications with forest datasets (Sec.~\ref{sec:perspectives}).

\section{Forest monitoring: current topics and challenges}
\label{sec:forest_topics_and_challenges}

%
Forest monitoring is an empirical science that is increasingly based on data-driven machine learning methods and, as such, benefits by improved data access through open data \cite{wulder_opening_2012, de_lima_making_2022}.
In particular, deep learning algorithms are widely recognized for their strong performance in diverse tasks, but their successful application often relies on large datasets to unleash their performance and enhance their generalization potential.
%
%
This section seeks to emphasize the importance of open access forest datasets for two primary purposes. First, to facilitate a more comprehensive exploration of current topics in the context of forest monitoring (Sec.~\ref{sec:forest_topics}). Second, to better assess forest monitoring related challenges (Sec. \ref{sec:forest_challenges}), in particular for machine learning practitioners. 

\subsection{Forest monitoring topics}
\label{sec:forest_topics}

Considering the significance of forests both economically and ecologically, forest monitoring encompasses a range of trackable forest attributes. 
Each of them can be sensed with different sensors, platforms and across different scales.  
The forest attribute itself (such as a regression of a biochemical property or the detection of tree individuals) together with the structure of the remotely sensed signals, 
collectively determine the appropriate machine learning algorithms to be employed.
These aspects are succinctly discussed in the following sections. 

\subsubsection{Forest extent and forest type mapping} 
\label{sec:topic_forest_map}
Tracking the extent of forests is crucial to understand the spatial distribution of forest resources, ecosystem services and assess the role of forest in land surface dynamics \cite{keenan_dynamics_2015}. Thereby, forest can be classified to different management, functional or ecosystems types (\textit{e.g.} coniferous, deciduous forests) \cite{zhang_glc_fcs30_2021, buchhorn_copernicus_2020}. In this regard, Earth observation data from long-term satellite missions (\textit{e.g.} Landsat or Sentinel described in Sec. \ref{sec:review_satellite}) enable to track forest extent dynamics across past decades \citep{hansen_high-resolution_2013}, enabling to assess conservation efforts and anthropogenic land cover change, such as deforestation for agricultural expansion \cite{curtis_classifying_2018}. 

\subsubsection{Tree species mapping} 
\label{sec:topic_species}
A fine-scaled representation of forest stands in terms of their species composition is relevant for forestry (\textit{e.g.} species-specific timber supply), biogeographical assessments (\textit{e.g.} climate change-induced shifts in species distributions) or biodiversity monitoring \citep{fassnacht_review_2016, wang_remote_2019, cavender-bares_remote_2020}. Recent developments in machine learning greatly advanced the identification of tree species in high resolution data (\textit{e.g.} imagery or LiDAR point clouds from drones and airplanes) using semantic and instance segmentation methods \citep{cloutier_influence_2023, schiefer_mapping_2020, li_ace_2022}. At large spatial scales, Earth observation satellite data, providing coarser spatial but high temporal and spectral resolutions, enable accurate assessments of tree species distributions using spatio-temporal machine learning methods \citep{ienco_combining_2019, bolyn_mapping_2022}.

\subsubsection{Biomass quantification} 
\label{sec:topic_biomass}
Forests provide cardinal ecosystem services through their provision of timber and their role as sinks in the terrestrial carbon cycle \citep{regnier_land--ocean_2022}. Tree biomass is primarily a product of the wood volume and density, while both properties are challenging to obtain from remote sensing data. Accurate biomass estimates of individuals trees can be obtained from close-range 3D representations acquired from terrestrial or drone-based LiDAR systems \citep{brede_non-destructive_2022}. More indirectly related information on crown and canopy structure derived from airborne or spaceborne LiDAR and SAR data can be used to estimate biomass at the stand scale \citep{le_toan_biomass_2011, lu_survey_2016}. Some studies have indicated the value of passive optical data from satellites, since forest biomass is partially correlated with foliage density \citep{besnard_mapping_2021, potapov_mapping_2021}. Given that precise large-scale biomass distributions can not be directly revealed through a single remote sensing modality alone, deep learning may play a crucial role to simultaneously exploit the suite and high dimensionality of available data modalities \citep{yang_new_2020}.

\subsubsection{Forest health, disturbance and mortality} 
\label{sec:topic_health}
In many regions, forest ecosystems are under pressure as globalization facilitates the introduction of exotic pests and pathogens, climate change exceeds the resilience and resistance of trees \citep{hartmann_climate_2022}, while nutrient and water cycles are affected by anthropogenic activities \citep{steffen_global_2005, trumbore_forest_2015}. A decline in tree health, \textit{e.g.} due to pathogen infestations or shortages of water and nutrients, can lead to a variety of symptoms, such as changing concentrations of multiple biochemical tissue properties (\textit{e.g.} pigments, carbohydrates, water content), which in turn can be sensed through multispectral or hyperspectral reflectance \citep{zarco-tejada_chlorophyll_2019, zarco-tejada_previsual_2018}. 
In this context, deep learning algorithms are very promising, due to their capability to exploit high dimensional data (\textit{e.g.} hyperspectral) and to translate it into a suite of foliage properties relevant to vegetation health \cite{cherif_spectra_2023}. 
An interconnected topic are globally increased rates of forest mortality \citep{hartmann_climate_2022, allen_global_2010}. In this context, a wealth of approaches was successfully employed at local scales, such as the detection of dead trees via semantic or instance segmentation techniques \citep{cloutier_influence_2023, sani-mohammed_instance_2022}, or at large-scales, such as the regression of annual cover of dead tree crowns in satellite image pixels with deep learning-based time series analysis \citep{schiefer_uav-based_2023}.

\subsubsection{Biophysical traits and functional ecosystem properties} 
\label{sec:topic_properties}
With accelerated biodiversity decline and environmental change, understanding  functional properties, their diversity across stands and landscapes, as well as their phenology (temporal dynamics), are essential to assess the resilience and resistance of ecosystems \citep{sakschewski_resilience_2016, thompson_forest_2009}. Given that trees through evolution developed different strategies to interact with light, their appearance studied with optical remote sensing signals can inform on a variety of functional traits, such as the foliage density, date of green up, or contents of different pigments, and carbohydrates \citep{schneider_mapping_2017, cherif_spectra_2023}. Such functional traits determine the configuration of an ecosystem and thereby modulate functional ecosystem processes \citep{gomarasca_leaf-level_2023, migliavacca_three_2021}, \textit{i.e.} fluxes of energy and matter between the terrestrial biosphere, pedosphere, hydrosphere, and atmosphere, including carbon, evapotranspiration, latent and sensible heat. Due to the cardinal importance of these fluxes in the Earth system, considerable efforts have been made to monitor them using a ground-based sensor network of flux towers (\textit{e.g.} FLUXNET \cite{baldocchi_fluxnet_2001}). Given the complexity of these ecosystem processes, deep learning is assumed to greatly enlarge our capabilities to exploit local flux towers and globally available remote sensing data to spatially and temporally extrapolate and understand forest ecosystem process \citep{jung_fluxcom_2019, elghawi_hybrid_2023, reichstein_deep_2019, campsvalls_deep_2021}.



\subsection{Forest monitoring challenges}
\label{sec:forest_challenges}

Forests are complex ecosystems dominated by trees. As living organisms, trees are affected by various abiotic and biotic factors, which influence their remotely sensed signal via their foliage properties and crown architecture \cite{kulawardhana_remote_2011}. Machine learning researchers wishing to develop algorithms to monitor forests using remote sensing data must be aware of these sources of biological variation and their origin. Because some of this biological variation is largely unpredictable but potentially clustered in space and time (\textit{e.g.} insect outbreaks affecting tree health, random genetic variation within tree species populations), it can be seen as a challenge as it might lead to systematic errors for prediction tasks. On the other hand, part of this variation is deterministic (\textit{e.g.} changes in leaf colour and other phenological changes driven by seasonal fluctuations that occur every year) and could be leveraged to improve model performances \cite{cloutier_influence_2023}. Another major, pervasive challenge in forest monitoring are the difficulties associated with the acquisition of ground data to train or validate machine learning models using remote sensing data. Below we summarise these primary challenges.

\subsubsection{Tree species} 
\label{sec:challenge_species}

There are an estimated 73,000 tree species on Earth \cite{cazzolla_gatti_number_2022}, the majority of which are found in the tropics. While tree species show many similarities (\textit{e.g.}, the presence of woody stems and branches), every tree species differs from one another in their chemical and structural make-up and how they will reflect solar radiation \cite{asner_functional_2014}. For example, tree foliage of different species comes in various shades of green that reflect the concentrations of pigments (\textit{e.g.} chlorophylls and carotenoids) \cite{gates_spectral_1965}. Likewise, tree species differ from each other in their leaf form crown structure \cite{verbeeck_time_2019}, which will affect the remotely sensed signal. Such foliar biophysical and crown structural variation among tree species is the result of millions of years of evolution and of adaptations to various environmental conditions \cite{meireles_leaf_2020}.

From a machine learning perspective, the biggest challenge associated with tree species diversity is that models trained on data from a given set of tree species might transfer poorly to other regions that host different species. However, the phylogeny and evolutionary distances of tree species are fairly well known \cite{zanne_three_2014}, and tree species that are closer phylogenetically tend to be more similar in their traits \cite{ackerly_conservatism_2009}. As such, phylogenetic correlations and distances among tree species can potentially be leveraged to improve model transferability.
Another interconnected challenge, elaborated upon in the following sections, pertains to the dynamic nature of tree species in relation to their leaf biophysical and crown structural characteristics.
Instead, each individual differs according to their abiotic (\textit{e.g.} microclimate, soil) and biotic environment (competition, herbivory) and as such the expression of foliage and crown properties can overlap between species \citep{fassnacht_review_2016}.

\subsubsection{Seasons and phenology}
\label{sec:challenge_phenology}

Trees are sessile organisms but they still respond dynamically to fluctuating seasons. In some cases, phenological properties, such as leaf onset, flowers or seeds, might be  of direct interest to monitor ecological phenomena or species \cite{wagner_flowering_2021}, in which case high-frequency multi-temporal imagery might be required. Indeed, phenological changes among species, for example changes in leaf color during autumn senescence, can help to distinguish tree species based on colour, which can be used to improve species classification models \cite{cloutier_influence_2023}. However, phenological properties may also hinder the transferability of models across time \citep{kattenborn_spatially_2022}. 
For instance, the information learnt by a machine learning model using data acquired in summer may not transfer to the same location in fall as trees may have changed in their leaf biochemical properties or the fraction of flowers and seeds in the canopy  \cite{schiefer_retrieval_2021}. 
In such cases, the temporal representativeness of data on individual tree species can be key \cite{kattenborn_spatially_2022}.

\subsubsection{Forest dynamics}
\label{sec:challenge_forest_dynamics}

The structure and composition of forests is strongly influenced by \emph{abiotic factors} such as climate, geology and soils, as well as water availability. For example, declining temperatures and/or growing season lengths with increasing latitude and/or elevation can filter out tree species that cannot tolerate low temperatures (\textit{e.g.} low frost resistance) or that do not have enough time to produce mature tissue once the growing season becomes too short \cite{korner_where_2016}. In addition, changes in soil nutrient availability driven by geomorphological processes can influence forest canopy biochemistry \cite{chadwick_landscape_2018}. Water supply is also important: too much water favours trees that can tolerate waterlogging, whereas too little water favours trees that can resist or recover from xylem cavitation \cite{choat_global_2012}. Much of these environmental influences on forest composition express themselves via tree species turnover; that is, changes in tree species composition across these spatial environmental gradients or discontinuities. However, changes in forest composition and structure can also arise through intraspecific variation. 
Applications of machine learning methods to forest monitoring should integrate these sources of variation.
In particular, incorporating environmental drivers of forest composition and structure as model inputs may help to transfer forest monitoring models from one region to the other.

Tree monitoring can also be affected by \emph{biotic factors} -- that is, by other organisms. First, pests and pathogens can impact tree health, foliage chemistry and/or water content, which in turn can affect the remotely sensed signal of forest canopies \cite{sapes_canopy_2022}. The health status of trees is often directly expressed via their foliage properties and crown architecture and therefore can cause a large variability in remote sensing signals \citep{zarco-tejada_previsual_2018, kattenborn2022anglecam}. In addition, the remotely sensed signals of trees can also be `overshadowed' by other organisms that live in their crowns (epiphytes), particularly in tropical environments \cite{baldeck_operational_2015}. Prominent examples are lianas or mistletoes.

Forest management activities as part of forest dynamics, such as harvesting, thinning, and pruning, can challenge the accurate mapping of forest attributes with remote sensing, as this crucial information is often unavailable but significantly impacts forest structure and composition. This lack of data therefore introduces uncertainty into remote sensing analyses and models.

\subsubsection{Data collection}
\label{sec:challenge_data}
As previously mentioned, forests can exhibit significant diversity in terms of their composition and structure across different locations and time periods due to a variety of factors.
This heterogeneity poses a particular difficulty in creating machine learning models for forest monitoring. Models developed for one region may not easily generalize to other regions that lie beyond the scope of the training data distribution.
One solution for this issue would involve training these models using extensive datasets that encompass the complete spectrum of conditions present in diverse forest environments.
Forest remote sensing data worldwide, especially acquired from sensors on satellites, are abundant and generally easily obtainable. In sharp contrast, there is a scarcity of ground-based data, including labels or annotations.

In contrast to other disciplines, annotating remote sensing data in the context of vegetation is often time consuming, costly, and complex. 
This phenomenon arises due to the fact that vegetation of various species or conditions frequently exhibit striking similarities, often referred to as `greenery'.
Moreover, vegetation communities often show smooth transitions across species or growth forms along environmental gradients. 
This aspect adds another layer of complexity to the task of distinguishing between individual plants, species, or growth forms \cite{kattenborn_review_2021}.
Often, field inventories become essential to validate annotations, such as the identification of tree species \cite{cloutier_influence_2023, kattenborn_convolutional_2020}, or when the properties of interest, such as stem diameters, cannot be directly extracted from remote sensing data and require on-site measurements conducted by human observers.
Gathering such field data is typically a time-intensive, expensive, and gradual process, leading to significant constraints on its accessibility.
Field data such as stem diameters are very important to estimate above-ground tree biomass because most published allometric equations use stem diameter as its primary predictor \cite{gonzalezakre_allodb_2022}.
Moreover, spatial coordinates frequently serve as the sole means of connecting field data to remote sensing data. However, GPS or GNSS geolocation in forest settings often introduces substantial uncertainties (ranging from meters to tens of meters), thereby posing challenges in accurately establishing a posteriori links between field observations and remote sensing data \cite{kattenborn_review_2021}.

Therefore, integrating various datasets is a strategy aimed at addressing the scarcity of annotated data, cutting down annotation expenses, and lessening the dependency on field-based ground truthing.
Nonetheless, this may result heterogenous datasets: In numerous cases, annotations vary (such as boxes, polygons, points), as well as their quality, across different applications.
Annotations are frequently customized to match particular remote sensing data characteristics, such as spatial resolution.
Hence, directly merging labels from different datasets is often not feasible, or at the very least, alternative approaches are necessary. One such approach is weakly supervised learning, where the potential lack of label quality is counteracted by leveraging a substantial quantity of data (see Sec.~\ref{sec:ml_weakly}).

The key takeaway from this section is that the development of machine learning models for forest monitoring will consistently involve a substantial surplus of unlabeled remote sensing data in comparison to labeled ground-truth data. This disparity arises due to the inherent challenges in obtaining labeled data.
This scenario is not exclusive to forest monitoring; rather, it is a prevalent aspect shared with other geospatial applications using remote sensing data \cite{rahaman_general_2022, mai_opportunities_2023} (see Sec.~\ref{sec:ml_foundation_models}).
This has two main implications for machine learning research aimed at improving forest monitoring. 
Firstly, there is a need to develop machine learning methods to forest monitoring that that can effectively utilize limited labeled data. One approach involves leveraging self-supervised learning techniques to extract valuable representations from the available data (see Sec.~\ref{sec:ml_ssl}).
Secondly, there exists a necessity for novel machine learning strategies, including active learning or alternative forms of model-assisted labeling. These approaches aim to expedite the process of label collection by human observers and reduce associated costs (see Sec.~\ref{sec:ml_active}).

\section{Machine learning perspectives and challenges}
\label{sec:ml_challenges}

Machine learning algorithms in computer vision have gained significant capabilities over the past decade, \textit{e.g.} in image classification \cite{krizhevsky_imagenet_2012, simonyan_very_2015, szegedy_going_2015, szegedy_rethinking_2016, he_deep_2016, szegedy_inception-v4_2017, hu_squeeze-and-excitation_2018, dosovitskiy_image_2021,liu_swin_2021, touvron_training_2021}, object detection \cite{ren_faster_2015, redmon_you_2016, redmon_yolov3_2018, liu_ssd_2016, he_mask_2017, li_mask_2023} and segmentation \cite{long_fully_2015, ronneberger_u-net_2015, lin_focal_2017, he_mask_2017, chen_encoder-decoder_2018, cheng_masked-attention_2022, kirillov_segment_2023}. 
%
While many successful algorithmic paradigms have been established, different applications differ widely across sensory modalities and domain-specific constraints, necessitating the adaptation of algorithms to fit specific needs.
For instance, detecting, localizing and segmenting objects in a scene have been explored for LiDAR point cloud \cite{yang_pixor_2018}, automotive RADAR  \cite{ouaknine_multi-view_2021} and medical echocardiography \cite{leclerc_deep_2019}.

Machine learning algorithms for remote sensing have been the subject of extensive innovation and application \cite{campsvalls_deep_2021, ma_deep_2019} for problems involving classification \cite{cheng_remote_2020, maxwell_implementation_2018}, object detection \cite{cheng_survey_2016, li_object_2020} and segmentation \cite{yuan_review_2021, hoeser_object_2020}.
%
In recent times, there has been a growing exploration of such techniques for forest monitoring purposes, aiming to enhance our understanding of forest composition, with a specific focus on tree species mapping (see Sec.~\ref{sec:challenge_species} and \ref{sec:topic_species}), \textit{i.e.} tree classification, tree detection, and tree segmentation \cite{fassnacht_review_2016, kattenborn_review_2021, diez_deep_2021, michalowska_review_2021}. These tasks are accomplished using modalities from diverse sensors to gather complementary information.

%
%
%
Nevertheless, the study of machine learning for forest monitoring has not received as much attention as autonomous driving or medical imagery, despite the importance of forest conservation, restoration, and management as natural solutions to the joint climate and biodiversity crises \cite{intergovernmental_panel_on_climate_change_ipcc_climate_2023}. Consequently, there are numerous unexplored machine learning challenges that need to be addressed in order to tackle climate change \cite{rolnick_tackling_2023}, including improving forest monitoring practices.
Can the challenges encountered in adapting machine learning for forest monitoring be beneficial in exploring the challenges in the field of biology and ecology?
This section will outline the current challenges in machine learning linked to forest monitoring, as described in Figure \ref{fig:challenges}, and discuss the diverse strategies employed to tackle them.

\subsection{Generalization} 
\label{sec:ml_generalization}

\emph{Generalization} in machine learning refers to the ability of an algorithm to continue to perform well when evaluated on data different from that it was trained on \cite{zhang_understanding_2017}. One may speak of both \emph{in-distribution} generalization (performance on data relatively similar to training data) and \emph{out-of-distribution} generalization (performance on highly different data). Out-of-distribution generalization can be especially relevant to forest-monitoring, since as mentioned in Section \ref{sec:review}, forest datasets have a wide range of variation in terms of geographical locations, species composition, sensors and scale (see Figure \ref{fig:scale}). Such variations introduce distinct data distribution shifts that need to be considered and addressed in forest monitoring tasks. For example, the effects of geographic variability of data have been examined in the context of tree species distributions \cite{f_dormann_methods_2007} and biomass estimation \citep{ploton2020spatial}. Simple algorithmic approaches to improve generalization include various forms of regularization \cite{zou_regularization_2005}, data augmentation \cite{shorten_survey_2019}, dropout \cite{srivastava_dropout_2014} and batch normalization \cite{ioffe_batch_2015}, while improving the breadth of training data, where possible, is also almost always beneficial in practice. However, generalization remains a very active field of research in machine learning. We consider two areas of work that may be of especial interest in forest-monitoring.
%

\subsubsection{Domain adaptation}
\label{sec:ml_domain}

\emph{Transfer learning} refers to transferring information learnt by a model on one set of problems to different set of problems \cite{weiss_survey_2016}. For example, one may pre-train a model on a large, commonly used dataset and then fine-tune it on a smaller dataset representing the specific problem in question.
Transfer learning can boost generalization when there is a significant distribution shift between training and inference \cite{csurka_domain_2017}. 
%
One particularly notable approach to transfer learning is \emph{domain adaptation}, where a model must be applied to target domains that are unknown or lacking labeled data \citep{soltani_transfer_2022}. Some domain adaptation approaches have already been applied in plant identification \cite{ganin_unsupervised_2015}.
%
%
Autonomous driving has witnessed significant exploration in the realm of \emph{unsupervised domain adaptation} (UDA), where a model is trained on labelled data from a source domain and unlabelled data from the target domain, with the objective of improving its performance specifically on the target domain.
It has been explored in the context of unlabelled or unseen source or target domains \cite{wilson_survey_2020} using generative \cite{hoffman_cycada_2018} or adversarial methods \cite{vu_advent_2019}. The UDA framework has also been explored for cross-modal learning considering domains as from different sensor modalities \cite{jaritz_xmuda_2020}.
Within the context of forest monitoring, this framework could prove particularly valuable for adapting a model from one forest to another, regardless of whether they belong to the same biome or not, to identify similar species across both the source and target domains (see Sec.~\ref{sec:topic_species} and ~\ref{sec:challenge_species}).
%
Additionally, this framework would be beneficial for adapting the model to address distribution shifts that occur between tree signature distributions (see Sec.~\ref{sec:topic_health}, ~\ref{sec:challenge_phenology} and ~\ref{sec:challenge_forest_dynamics}) as well as between different sensors (see Sec.~\ref{sec:challenge_data}).
%
Domain adaptation has been investigated in the field of remote sensing mostly in the context of extrapolation across time or geographical region, including approaches for both aerial and satellite data \cite{wang_exploring_2022, shi_unsupervised_2022, xu_universal_2023, ma_unsupervised_2023, arnaudo_hierarchical_2023} .
Such work holds potential for training generalizable algorithms for forest monitoring, such as adapting models from PhenoCams to satellite images \cite{kosmala_integrating_2018}.

\subsubsection{Foundation models}
\label{sec:ml_foundation_models}
\emph{Foundation models} are models that can operate on diverse sets of input modalities, scales, data regimes and downstream tasks.
They refer to large-scale multi-modal and multi-task models, which have opened up research in generalization capacities such as increasing performances in applications unseen during training \cite{bommasani_opportunities_2021}. 
Most of the discussed machine learning strategies can be further explored by training foundation models with diverse datasets, thereby enhancing their generalization capabilities. 
Forest datasets encompass a wide range of scales, ranging from field measurements to estimated world maps (refer to Figure \ref{fig:scale}), as well as varying resolutions and modalities for different tasks (as outlined in Section \ref{sec:review}). The data diversity necessitates the utilization of generalized deep learning architectures.
%
Influenced by the success of large language models (LLMs) \cite{radford_language_2019, devlin_bert_2019, brown_language_2020, chowdhery_palm_2022, hoffmann_training_2022, radford_learning_2021, touvron_llama_2023, driess_palm-e_2023}, recent advancements in computer vision have led to the development of models that incorporate multiple modalities and can perform multiple tasks simultaneously. 
In recent studies, researchers have explored the concept of multi-task vision by utilizing natural images \cite{cheng_per-pixel_2021, cheng_masked-attention_2022, li_mask_2023, kirillov_segment_2023} or by employing text to enhance performance in vision-based tasks \cite{dancette_dynamic_2022, xu_groupvit_2022, jain_oneformer_2023}. 
In the realm of integrating image and text for performing tasks on both modalities, alternative approaches have been developed to improve performances by benefiting from their combination \cite{zhu_uni-perceiver_2022, li_uni-perceiver_2023}.
Additionally, generalist models have been constructed to be agnostic to specific modalities and tasks \cite{jaegle_perceiver_2021, jaegle_perceiver_2022}, enabling them to handle diverse modalities and tasks with a unified approach.
%
In the field of computer vision, for example, the Segment Anything Model (SAM) \cite{kirillov_segment_2023} has demonstrated the capability to perform instance segmentation by leveraging prompts in conjunction with input images. 
These architecture frameworks hold significant value for forest monitoring tasks, enabling the detection, segmentation, and estimation of tree properties over large geographical areas, such as their canopy surface or their above-ground biomass \cite{tucker_sub-continental-scale_2023, tolan_sub-meter_2023}.

The utilization of foundation models with remote sensing data is still in its infancy. 
However, promising advances have been made in developing multi-modal architectures \cite{zhang_mmformer_2023} and temporal-based approaches \cite{garnot_panoptic_2021, garnot_multi-modal_2021, tarasiou_vits_2023} specifically tailored for precise tasks in remote sensing applications. 
Based on the masked autoencoder (MAE) pretraining method \cite{he_masked_2022}, multi-modal and multi-task architectures have been developed for Earth observation applications, in particular for land use and land cover (LULC) estimation \cite{sun_ringmo_2022, cong_satmae_2022, reed_scale-mae_2022, tseng_lightweight_2023}.
%
%
These architectures address the challenges posed by data collected from sensors that record diverse physical measurements, such as multispectral or SAR data in remote sensing \cite{reed_scale-mae_2022, yamazaki_aerialformer_2023, pan_multi-scale_2023}, as well as in natural images \cite{themyr_full_2023}. While different spectral, spatial and temporal resolutions have been considered in previous works, there remains a lack of exploration regarding the resolution gap between datasets captured by aerial and satellite sensors.
%
The integration of multi-modal, multi-task, and multi-scale architectures is expected to significantly enhance the generalization capabilities of models for forest monitoring tasks at global scale (see Sec.~\ref{sec:challenge_data}). 
By training these algorithms with various type of datasets and specialising them for forest monitoring tasks, they could effectively
deliver improved performance across different geographical regions such as for species cover estimation or above-ground biomass estimation (see Sec.~\ref{sec:topic_biomass}).
%

\subsection{Learning from limited data}
\label{sec:ml_limited_data}

There are a growing number of massive datasets and algorithms leveraging them, including across remote sensing \cite{bastani_satlas_2022, sumbul_bigearthnet-mm_2021,rahaman_general_2022, mai_opportunities_2023}. 
However, many of the most powerful machine learning approaches are \emph{supervised}, and therefore require labels, which can be challenging, time-consuming, and costly to obtain.
There has been considerable attention given to the problem of learning from limited labeled data; we here present several families of approaches and their relevance to forest-monitoring.

\subsubsection{Self-supervised learning}
\label{sec:ml_ssl}
Situated in-between supervised and unsupervised learning, the \emph{self-supervised learning} paradigm involves training a model to reconstruct certain known relationships between or within the datapoints themselves. The resulting model can then be fine-tuned with actual labeled data or directly applied to solve the downstream task.
%
%
Self-supervised approaches in computer vision include discriminative approaches that distinguish between positive and negative samples while separating their  representation (\textit{e.g.} contrastive learning) \cite{gidaris_unsupervised_2018, he_momentum_2020, chen_simple_2020, caron_emerging_2021, oquab_dinov2_2023}, and generative approaches learning representation by reconstruction \cite{lehtinen_noise2noise_2018, he_masked_2022}.
%
The utilization of self-supervised learning in remote sensing has experienced significant growth due to the abundance of unlabelled open access data \cite{tao_self-supervised_2023}.
For instance, geolocation of satellite images have been exploited with a contrastive approach \cite{mai_csp_2023,ayush_geography-aware_2021}. 
Multi-spectral and SAR data have been reconstructed based on the temporal information \cite{cong_satmae_2022, yadav_unsupervised_2022}, for multi-scale reconstruction \cite{reed_scale-mae_2022} and for denoising \cite{dalsasso_sar2sar_2021, dalsasso_as_2022, meraoumia_multitemporal_2023}.
Emerging cross-modal approaches, encompassing both discriminative \cite{jain_multimodal_2022} and generative \cite{jain_multi-modal_2023} techniques, are being developed to harness the complementary nature of aligned samples.
Self-supervised learning will greatly unleash the potential of remote sensing data in the area of forests \cite{ge_novel_2023} by learning textural and geometrical structures of forests and trees without labels (see Sec.~\ref{sec:challenge_species} and ~\ref{sec:challenge_data}).

\subsubsection{Weakly-supervised learning}
\label{sec:ml_weakly}
Obtaining precise and detailed annotations, \textit{e.g.} for tree crown instance segmentation, can be both costly and time-consuming. 
Although self-supervised learning aims to learn representations from pretext tasks, it still necessitates precise annotations for fine-tuning the model in a downstream task. 
In cases where precise annotations are not available, coarse-grained and potentially inaccurate annotations, or even single point locations, can be utilized as weak labels in \emph{weakly-supervised learning} approaches \cite{zhou_brief_2018}.
%
Given their cost-effectiveness and efficiency, computer vision methods have been developed to leverage weak annotations while addressing their inherent inaccuracies \cite{zhou_brief_2018}.
Weakly-supervised learning has been explored in the realm of object location \cite{oquab_is_2015}, object relationship estimation \cite{peyre_weakly-supervised_2017}, instance segmentation \cite{ahn_weakly_2019} and contrastive learning \cite{zheng_weakly_2021}.
%
Obtaining high-quality annotations for remote sensing data is particularly difficult due to their poor spatial resolution or the physics of the sensors used. Weakly-supervised learning has therefore been investigated for Earth observation tasks 
including object detection \cite{dingwen_zhang_weakly_2015, han_object_2015, yao_automatic_2021}, LULC semantic segmentation \cite{wang_weakly_2020, yao_semantic_2016} and plant traits regression \cite{cherif_spectra_2023, schiller_deep_2021}.
%
Recently, weakly-supervised methods have been investigated in the areas of tree classification \cite{illarionova_tree_2021}, tree counting \cite{amirkolaee_treeformer_2023}, tree detection \cite{aygunes_weakly_2021}, and segmentation \cite{gazzea_tree_2022} using multispectral data (see Sec.~\ref{sec:challenge_data}).

\subsubsection{Active learning}
\label{sec:ml_active}
Even highly precise and fine-grained annotations are generally less useful if present in only small quantities. 
To address this limitation, \emph{active learning} strategies have been developed to identify and select the optimal way to select a small set of training datapoints to label \cite{cohn_active_1996}.
%
These strategies for sample selection often rely on estimating the uncertainty of a model \cite{gal_deep_2017}, for instance using variational approaches \cite{sinha_variational_2019} or estimated with a loss function \cite{yoo_learning_2019}.
They have demonstrated their effectiveness in scenarios where the amount of labeled data is limited, particularly in the context of image classification \cite{gal_deep_2017, sinha_variational_2019, yoo_learning_2019}, object detection \cite{roy_deep_2019} and semantic segmentation \cite{siddiqui_viewal_2020}.
%
Active learning has also been investigated for remote sensing applications, including classification \cite{tuia_survey_2011}, object detection \cite{qu_deep_2020} and LULC semantic segmentation with hyperspectral data \cite{li_semisupervised_2010, li_hyperspectral_2011, zhang_active_2016}.
%
Its application would be helpful for forest monitoring to optimize and create relevant human annotations (see Sec.~\ref{sec:challenge_data}).

\subsubsection{Few-shot learning}
\label{sec:ml_fewshot}
Yet another approach to limited data availability is \emph{few-shot learning}, which refers to efficient fine-tuning of a pretrained model using only a few annotated datapoints.
%
Few-shot learning has been approached from different perspectives, considering the comparison between the small annotated dataset and the data used for pretraining the model -- for example, by quantifying the similarities between these datasets \cite{vinyals_matching_2016}, constructing mixtures of feature embeddings \cite{snell_prototypical_2017} or adapting the optimization scheme through meta-learning \cite{finn_model-agnostic_2017}.
%
Motivated by the limited availability of annotations, applications of few-shot learning in remote sensing tasks have been investigated. 
For instance, methods based on feature similarity \cite{zhang_few-shot_2020, alajaji_few-shot_2020, alosaimi_self-supervised_2023} and metric learning \cite{liu_deep_2019}, aiming at separating representations in an embedding space, have been explored for LULC classification with either multispectral or hyperspectral data.
Objects have also been detected by learning meta features \cite{deng_few-shot_2022}.
%
Metric learning techniques have also been utilized in the context of few-shot learning for semantic segmentation tasks \cite{jiang_few-shot_2022} or meta learning with multispectral and SAR data \cite{ruswurm_meta-learning_2020}.
%
Few-shot learning has been explored for tree species classification using feature similarity \cite{chen_new_2021} and would be beneficial to recognize a species or estimate the characteristics of a tree with minimal manual annotations (see Sec.~\ref{sec:challenge_phenology} and ~\ref{sec:challenge_data}).

\subsubsection{Zero-shot learning}
\label{sec:ml_zeroshot}
The machine learning community has also shown interest in developing methods for training algorithms to differentiate unseen classes without any explicitly annotated samples at all, which is known as \emph{zero-shot learning} \cite{xian_zero-shot_2018}.
%
In order to categorize unseen classes, the task of zero-shot learning has been accomplished by projecting image and word embeddings \cite{socher_zero-shot_2013} or known semantic attributes \cite{lampert_attribute-based_2014} into a shared space.
%
Zero-shot learning has also been investigated by incorporating a mixture of embeddings from the source domain before computing similarities with the target domain, which includes the unseen classes \cite{zhang_zero-shot_2015}.
Generative approaches have been developed to create visual feature embeddings of unseen classes from word embeddings for zero-shot semantic segmentation \cite{bucher_zero-shot_2019}.
%
Zero-shot learning has also garnered attention in remote sensing applications, including
 combining multispectral data and word embeddings for classification tasks \cite{li_zero-shot_2017, li_robust_2021}
and initial exploration of applying zero-shot learning to classify hyperspectral data \cite{freitas_hyperspectral_2022}.
Generative approaches have also been used with remote sensing data to create visual embeddings from word embeddings  \cite{li_generative_2022}.
%
Zero-shot learning presents a promising approach for forest monitoring, enabling the adaptation of models in regions where previously unseen species are encountered (see Sec.~\ref{sec:challenge_phenology} and ~\ref{sec:challenge_data}).
By leveraging tree taxonomy hierarchy and meta characteristics to align with visual embeddings \cite{sumbul_fine-grained_2018}, a vast research potential emerges. 
Notably, utilizing foundation models that have demonstrated strong zero-shot learning capabilities \cite{brown_language_2020, radford_learning_2021} further enhances this potential.
In the following paragraph 
will explore methods concerning domain-specific objectives, focusing on the consideration of physical and biological constraints and their applications.

\subsection{Domain-specific objectives}
\label{sec:ml_metrics}

Machine learning methods commonly use a fairly limited set of metrics to evaluate success, such as (macro or micro) accuracy of labels and cross-entropy loss for classification tasks, mean squared error or mean average error for regression tasks, etc. However, these uniform metrics do not necessarily reflect the realities of real-world use cases  \cite{birhane2022values}, where criteria for success may be much more nuanced or domain-specific. In this section, we consider two other families of objectives that may frequently be of relevance in forest-monitoring.
%
%

\subsubsection{Constraints on data}
\label{sec:ml_constraints}

Depending on the domain of application, the outputs of a machine learning pipeline may have specific constraints that must be satisfied if the answer is to be useful or even possible. For example, climate variables may need to obey physical laws such as conservation of energy, engineered systems may need to obey the laws of mechanics, etc. Machine learning models to work with such variables have increasingly been designed with soft constraints \cite{harder_physics-informed_2022,ouaknine_multi-view_2021}, which impose penalties for constraint violation, or hard constraints \cite{donti_dc3_2021,geiss_strict_2021,harder_physics-constrained_2023}, where the constraints are strictly enforced by the design of the algorithm.
Compared to physics- and engineered-based constraints, fewer authors have to date integrated biological constraints into ML algorithms. 
Dynamics of biological systems have been included in a deep learning optimization scheme as hard constraints from ordinary differential equations \cite{yazdani_systems_2020}.
There are potential opportunities for incorporating biological constraints in forest monitoring by considering phenological \cite{richardson_intercomparison_2018} or biophysical traits, or ecosystem properties (see Sec.~\ref{sec:topic_properties}), for example by considering the ratio of tree height and canopy size. These constraints could be particularly valuable in tasks such as semantic segmentation or biomass estimation.

Domain-specific constraints on data may also pose opportunities for improving the design of machine learning models. 
The design of deep learning model architectures can incorporate considerations for, or reconstruction of, physical properties. 
For instance, a physics-informed architecture has been developed for super-resolution in turbulent flows, incorporating partial differential equations as a form of regularization \cite{jiang_meshfreeflownet_2020}. 
Similarly, RADAR-based architectures have been created to reconstruct physical properties for scene understanding in the context of autonomous driving \cite{ouaknine_multi-view_2021, rebut_raw_2022}.
%
Leveraging the properties of multiple sensors has also been employed to fuse their representations \cite{ouaknine_deep_2022} or to generate annotations from one modality to another \cite{ouaknine_carrada_2021, schiefer_uav-based_2023}.
In remote sensing, self-supervised learning has benefited from SAR physical properties by considering a pretext denoising task \cite{dalsasso_sar2sar_2021, meraoumia_multitemporal_2023}, or by separating and reconstructing the real from the imaginary part of the signal \cite{dalsasso_as_2022}.
Such methods could also be explored by exploiting various sensors to learn representation of forests and trees (see Sec.~\ref{sec:challenge_data}).

%


%
%
%

\subsubsection{Uncertainty quantification}
\label{sec:ml_uncertainty}
Biological phenomena adhere to intricate rules that are challenging to estimate and often exhibit inherent uncertainties. 
The estimation of prediction uncertainty aids in obtaining a better understanding of the strengths and limitations of a machine learning model.
%
%
The overall uncertainty of these models comprises both aleatoric and epistemic uncertainties \cite{gal_uncertainty_2016}.
They both can be distinguished based on their origins. 
Aleatoric uncertainty arises from the inherent noise present in the data and label distributions, while epistemic uncertainty is associated with the model itself, encompassing its estimated parameters and structural characteristics. 
%
Approaches have been devised to estimate the uncertainties of deep neural networks, \textit{e.g.} by using a Bayesian approach such as Monte Carlo dropout \cite{gal_dropout_2016}, by using adversarial training combined with model ensembles \cite{lakshminarayanan_simple_2017}, by predicting the uncertainty distribution \cite{malinin_predictive_2018} or by learning an auxiliary confidence score from the data \cite{corbiere_addressing_2019, corbiere_robust_2022}.
%
Similar methods have been applied to estimate uncertainty in remote sensing data for crop yield estimation \cite{ma_corn_2021} or for road segmentation \cite{haas_uncertainty_2021}.
%
The quantification of uncertainties in forest monitoring methods has been carried out to assess both aleatoric and epistemic uncertainties (see Sec.~\ref{sec:forest_challenges}). This is commonly performed to evaluate the uncertainty of predictions on large-scale maps, utilizing low-resolution satellite data.
The uncertainty of plant functional type has been studied for classification in Siberia \cite{ottle_use_2013}.
%
Estimating the uncertainty of above ground biomass has also been conducted to establish a range of estimated values in carbon stock maps \cite{patterson_statistical_2019, santoro_global_2021} (see Sec.~\ref{sec:topic_biomass}).
%
To quantify uncertainty, these methods utilize standard deviations or output probabilities of the model.
Recent studies have taken a step further in estimating tree carbon stocks in semi-arid sub-Saharan Africa north of the Equator by combining uncertainty from both allometric equations and predicted crown segmentation, utilizing field measurements \cite{tucker_sub-continental-scale_2023}. 
%
There has been limited application of advanced uncertainty quantification methods, whether associated with the data or the predictive model, in the context of forest monitoring.

Despite the extensive application of the presented machine learning techniques in remote sensing, their utilization for forest monitoring has been relatively limited. 
This presents numerous opportunities to gain deeper insights into the composition of forests while achieving generalization at a large scale. 
However, it is crucial to have access to high-quality, diverse, and sufficient datasets in order to effectively explore machine learning strategies. 
In the following section, we will review open access forest datasets, providing information on their size, tasks, scale, and modalities.

\section[Review of open access forest datasets]{Review of open access forest datasets}
\label{sec:review}

Open access datasets are essential to drive the scientific community in general to exploring forest biology challenges, in particular by using machine learning strategies (see Sec. \ref{sec:ml_challenges}).
Deep learning algorithms have demonstrated strong performance in various forest monitoring tasks, such as tree classification or segmentation \cite{kattenborn_review_2021}.
The availability of open access datasets has played a significant role in enhancing the algorithm performances and expanding their applications on a larger scale.
In this particular field, the use of data, from the tree to the country level (see Fig. \ref{fig:scale}), distributed in the entire globe, must be taken into consideration.
Algorithms have been trained for forest monitoring by leveraging datasets that encompass different scales, modalities, and tasks \cite{guimaraes_forestry_2020, kattenborn_review_2021, michalowska_review_2021}.
However, the limited availability of data sources often restricts public access, thereby impeding the progress of extended research projects.
While the scientific community emphasizes the importance of reproducible experiments, it is worth noting that some datasets do not fully adhere to the fair principles\footnote{\url{https://www.go-fair.org/fair-principles/}}, which encompass aspects like documentation and findability.

\begin{figure}[t]%
\includegraphics[width=1.0\textwidth]{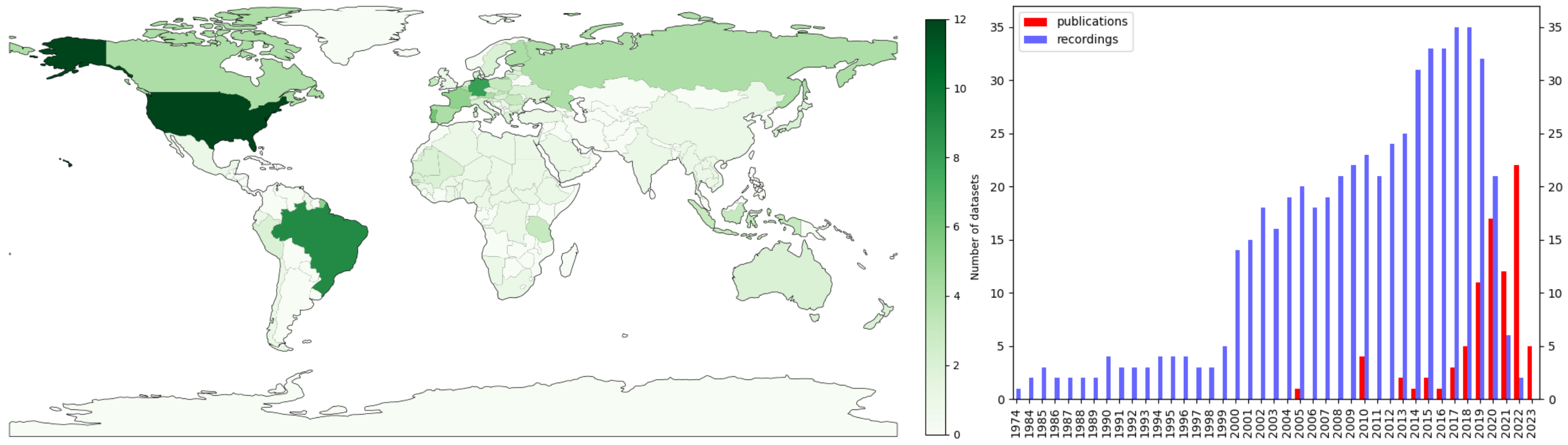}
\caption{
\textbf{Distribution of the reviewed open access forest datasets.} (Left) World map of the location of the reviewed datasets at the country level. Most of the datasets are regional and do not reflect the entire associated country. The datasets categorized with a `Worldwide' location or at the continent level have been excluded for visualization purposes. (Right) Distributions of the publication years and recording years used and / or released in the associated datasets}
\label{fig:distributions}
\end{figure}

While there is still a considerable quantity of publicly available datasets, it is important to acknowledge that they may have certain limitations that restrict their impact in machine learning applications for forest composition analysis. These limitations can include factors such as the size of the dataset or the specific type of data that is released.
This section aims to review forest monitoring datasets considering the following criteria:
\begin{enumerate}
    \item The dataset should be open access, \textit{i.e.} without any request requirement;
    \item The dataset should be related to at least one published article, exceptions have been made for datasets that are available as preprints, but are considered to be must-see datasets;
    \item The dataset should be focused on the composition of the forest, excluding event-based specific ones (\textit{i.e.} wildfire detection);
    \item A land use and/or land cover (LULC) dataset should contain more than a single plant functional type (\textit{i.e.} conifers or deciduous) since a focus is made on better understanding the composition of the forest;
    \item The dataset should be at the tree level at least, excluding datasets at the organ or cellular level considered as out of the scope of this review (\textit{e.g.} leaf spectra or root scans);
    \item The dataset should contained at least $O(1000)$ trees. 
\end{enumerate}
%
Based on these criteria, 86 datasets have been identified representing a wide range of geographical locations and spanning from 1974 to 2022. The datasets are associated with publications from 2005 to 2023, as depicted in Figure \ref{fig:distributions}.

The scope of the presented review is broad, it is likely that other datasets meeting these requirements have been missed.
Based on this motivation, the study is supported by \textbf{OpenForest}\footnote{The catalogue contains all urls to access the datasets which are not included in this article to ensure a temporal consistency. \textbf{OpenForest} is available here: \url{https://github.com/RolnickLab/OpenForest}}, a dynamic catalogue integrating the reviewed datasets and open to updates from the community. Updates on \textbf{OpenForest} will be restricted with the criteria detailed above. We hope to motivate researchers by grouping our efforts to create the largest database of open access forest datasets and thus create synergies 
on forest monitoring applications.

This section will review open access forest datasets grouped at different scales as presented in Figure \ref{fig:scale}: inventories (Sec. \ref{sec:review_inventories}), ground-based recordings (Sec. \ref{sec:review_ground}), aerial recordings (Sec. \ref{sec:review_aerial}), satellite recordings (Sec. \ref{sec:review_satellite}) and country or world maps (Sec. \ref{sec:review_maps}). Datasets composed of mixed scales are finally presented (Sec. \ref{sec:review_mixed}). 

Each section will detail the overall scope of the presented datasets with the specificity of the sensors used to record the data, the information related to each dataset and their applications.
In each section, the reviewed datasets will be categorized in tables respectively to the scale of the released data. 
In these tables, the publication and recording years are differentiated to better understand the temporal scope of the datasets. The recordings years are distinguished with a new line while time series are represented by an upper dash.
Each table will relate the available modalities in the `data' column. This one is separated with the `spatial resolution', or `spatial precision' columns (except for inventories) with a dashed line to associate a resolution to the corresponding modality.
Each section will also discuss the limits of current open access datasets to motivate our perspectives presented in Section \ref{sec:perspectives}.
The following section will review inventory datasets as the smallest scale of recordings that have been taken into account.

\subsection{Inventories}
\label{sec:review_inventories}

Historically, forests have been mostly locally or regionally inventoried based on stratified plot samples acquired in the field \cite{jucker_tallo_2022}.
Digitized and open access inventories generally cover small areas, consisting of dozens or a few hundred trees, which limits their impact on the machine learning community (Sec. \ref{sec:ml_challenges}). 
As defined in Section \ref{sec:review}, this section is focused on medium to large scale inventories with at least $O(1000)$ trees.
A significant part of reviewed inventory datasets are mixed with modalities at different scales, which will be detailed in Section \ref{sec:review_mixed}.

Inventory datasets are summarized in Table \ref{tab:inventories}, the size of the datasets is quantified by the number of trees. 
Inventory datasets are composed of various measurements. They commonly contain tree height, canopy diameter, diameter at breast height (DBH) or diameter at soil height (DSH) \cite{national_ecological_observatory_network_neon_vegetation_2023, gastauer_tree_2015, perez-luque_dataset_2015, perez-luque_land-use_2021, oliveira_structure_2017, jucker_tallo_2022}. In specific cases, wood density, bark density and bark thickness are also measured \cite{schepaschenko_forest_2019, farias_dataset_2020, kindermann_dataset_2022}. These information are particularly useful to estimate the tree density, the above ground biomass (AGB) or the tree carbon stock at large scale \cite{tucker_sub-continental-scale_2023} even if the inventories have not been released with the estimated maps \cite{patterson_statistical_2019, dionizio_carbon_2020}.

Species, genus and family of the trees are generally provided. 
This hierarchy of labels coming alongside with the tree geo-location make inventories a very accurate datasets for understanding forest composition.
However, they are geographically sparse and centered in a specific location to reduce measurement efforts \cite{laar_forest_2007, motz_sampling_2010}. 
As an exception, Tallo \cite{jucker_tallo_2022} groups inventories from all around the world with an unprecedented number of species reported. The latter could have an impact on estimating tree species distribution at large scale.

Considering that inventories contain annotations of trees or tree clusters, they open possibilities to segment tree canopies according to their taxonomic levels, regress continuous metrics (\textit{i.e.} height, biomass) or even locate tree individuals by predicting their coordinates or crown perimeter \cite{tucker_sub-continental-scale_2023}.
Another example could be to estimate the wood density of a tree or its carbon stock using allometric equations with information on taxonomy and height measured on the field \cite{zianis_biomass_2005}. 
Inventories could also be combined with other modalities and used as annotations for larger scale tasks.
As an example, remote sensing datasets presented in the following sections in the same geographic locations could be associated to inventories to enhance the precision of their annotations. 
While establishing this connection between ground measurements and remote sensing data presents its own set of challenges.
In the following section, we will review datasets of ground-based recordings.

\begin{table*}[ht]
\fontsize{6.5pt}{7.5pt}\selectfont 
\renewcommand{\arraystretch}{1.5} 
\setlength\tabcolsep{5pt} 
\caption{Review of open access forest inventories datasets}
{\begin{fntable}
\centering
\begin{tabular}{p{1.3cm} | p{0.4cm} | p{1cm} | p{1cm} | p{1.4cm} | p{0.4cm} | p{1cm} | p{1cm} | p{1cm} | p{1.1cm}}
\toprule
Dataset  & Publi. year & Recording year & Dataset size & Data & Time series & Potential task(s) & \#Classes & Location  & License  \\
\midrule

NEON Vegetation Structure \cite{kampe_neon_2010, national_ecological_observatory_network_neon_vegetation_2023} & 2010 & 2014-2021 & Unknown & Location \newline Height \newline Crown diam. \newline Stem diam. \newline Health & Yes & OL \newline Classif. \newline Reg. & 2826 species \newline 949 genus \newline 316 families & USA & CC0-1.0 \\

\rowcolor{lightgray}
Seu Nico Forest	\cite{gastauer_tree_2015} & 2015 & 2001-2010 & 2868 trees & Location \newline DBH \newline Height \newline Soil & Yes & OL \newline Classif. \newline Reg. & 228 species \newline 139 genera \newline 54 families & Brazil & CC0-1.0	\\

MIGRAME	\cite{perez-luque_dataset_2015, perez-luque_land-use_2021} & 2015 & 2012-2014 & 3839 trees & Location \newline Diameter \newline Height & No & OL \newline Classif. \newline Reg. & 6 species \newline 5 genus \newline 5 families & Spain & CC-BY-NC-4.0 \\

\rowcolor{lightgray}
Northern Brazilian Amazonia	\cite{oliveira_structure_2017} & 2017 & 2014 \newline 2015 & 1026 trees & Location \newline DBH \newline DSH & No & OL \newline Classif. \newline Reg. & 52 species \newline 28 families & Brazil & CC-BY-4.0 \\

Forest Observation System (FOS) \cite{schepaschenko_forest_2019} & 2019 & between 1999 and 2018 & 1646 trees & Location \newline Canopy height \newline Tree density \newline Wood density \newline AGB & No & OL \newline Reg. & N/A & Worldwide & CC-BY-4.0 \\ 

\rowcolor{lightgray}
SeasonWatch	\cite{ramaswami_using_2020} & 2020 & 2011-2019 & 352K trees & Location \newline Estimation of leaf, flower and fruit quantity & Yes & OL \newline Classif. \newline Reg.  & 136 & India & CC-BY-4.0 \\

Maraca Ecological Station	\cite{farias_dataset_2020} & 2020 & 2018 \newline 2019 & 680 trees & Location \newline Bark thickness \newline Bark density \newline Wood density & No & OL \newline Classif. \newline Reg. & 110 species \newline 40 families & Brazil & CC-BY-4.0 \\

\rowcolor{lightgray}
Tallo \cite{jucker_tallo_2022} & 2022 & Unknown & 499K trees & Location \newline Diameter \newline Height \newline Crown radius & No & OL \newline Classif. \newline Reg. & 5163 species \newline 1453 genus \newline 187 families & Worldwide & CC-BY-4.0 \\

African Savanna	\cite{kindermann_dataset_2022} & 2022 & 2018 \newline 2019 & 6179 trees and shrub & Location \newline Height \newline Stem circ. \newline Canopy diam. \newline Wood density & No & OL \newline Classif. \newline Reg. & 65 species \newline 6 categ. & Namibia & CC-BY-4.0 \\

\bottomrule

\end{tabular}
\footnotetext[]{{Acronyms}: \textbf{N/A}: non applicable; \textbf{Unknown}: non provided by the authors; \textbf{DBH}: diameter at breast height, \textbf{DSH}: diameter at soil height; \textbf{AGB}: above ground biomass; \textbf{OL}: object localization; \textbf{Reg.}: regression; \textbf{Classif.}: classification. Note that the dataset size measured in \textbf{K} are $O(10^3)$.}
\end{fntable}}
\label{tab:inventories}
\end{table*}

\subsection{Ground-based recordings}
\label{sec:review_ground}



The fine-scaled composition of forests can be understood by visualising the trees within or under their canopy. 
Ground-based datasets are composed of recordings inside the forests, under the tree canopy. 
Trunks and small trees, invisible from a bird's eye view, can be captured with cameras recording red-green-blue (RGB) images per example. These data are sometimes recorded in time series \textit{e.g.} PhenoCams \cite{klosterman_evaluating_2014, brown_using_2016}. 
The use of data recorded by sensors by machine learning algorithms help to have a broader context and more tree information in the samples compared to inventories.

Ground-based datasets are reviewed in Table \ref{tab:ground}. The dataset size has been measured in hectares (ha) corresponding to the studied surface, in number of trees in the area or in number of samples, which may differ between synthetic or real samples \cite{grondin_tree_2022}.

Stereo cameras are parameterized to estimate the depth of a scene differentiating trees and objects from the background in the forest \cite{grondin_tree_2022}. 
Thermal cameras have also been used to record trees' signature \cite{still_thermal_2019} and distinguish them from other objects \cite{da_silva_visible_2021, da_silva_unimodal_2021, da_silva_edge_2022, reis_forest_2020}.
In specific cases, camera images have been annotated with bounding boxes around trees to detect them \cite{tremblay_automatic_2020, grondin_tree_2022}.
Only two reviewed datasets located in Canada have been annotated with several species classes to combine detection and classification of trees \cite{tremblay_automatic_2020, grondin_tree_2022}.
Since these datasets also provide inertial measurement unit (IMU), a potential task could be to predict the next move of an automated agent in a forest.

Forest geometry is also being intensively studied from the ground by using LiDAR - typically referred to as terrestrial laser scanning (TLS). This active sensor records 3-dimensional scenes with photon reflections and can be applied from tripods or be combined with IMUs to enable mobile laser scanning. 
It is not impacted by sun lighting conditions and well suited to understand the structure of forests and trees such as measuring, gap fraction, stand density, tree height, DBH, volume or biomass \cite{hackenberg_simpletree_2015, liang_terrestrial_2016, tremblay_automatic_2020}. The spatial resolution of ground-based LiDAR recordings are either expressed in the averaged number points per meter squared, or in the precision of localisation of each point, based on information provided by the authors.
The generated LiDAR point clouds have been used for instance segmentation \cite{burt_extracting_2018, tremblay_automatic_2020, grondin_tree_2022}, \textit{i.e.} segment each tree independently and associate them an identification number, or key-point detection, \textit{i.e.} localizing points of interest for each tree. 
%

Ground-based datasets are useful to understand the composition of forests under the tree canopy and recordings were difficult to automatize until recently \cite{calders_strucnet_2023}.
Literature lacks large-scale annotated datasets although they can provide information at high spatial and temporal resolution and from perspectives that aerial and satellite recordings cannot. 
Providing both ground-based and aerial-based recordings \cite{soltani_transfer_2022} informing both above and below tree canopy would facilitate transfer and bridging machine learning applications between different modality scales (for details see Section \ref{sec:perspectives}).
The next section will review aerial recordings datasets.

\begin{table*}[t!]
\fontsize{6.5pt}{7.5pt}\selectfont 
\renewcommand{\arraystretch}{1.5} 
\setlength\tabcolsep{5pt} 
\caption{Review of open access ground-based forest datasets}
{\begin{fntable}
\centering
\begin{tabular}{p{1.5cm} | p{0.4cm} | p{1cm} | p{1.3cm} | p{1.2cm} : p{1cm} | p{0.4cm} | p{0.7cm} | p{0.8cm} | p{1cm} | p{1.1cm}}
\toprule
Dataset  & Publi. year & Recording year & Dataset size & Data  & Spatial precision & Time series & Potential task(s) & \#Classes & Location  & License  \\
\midrule

NOU-11 / KARA-001 \cite{burt_extracting_2018} & 2018 & 2015 & 1 ha and 425 trees / \newline 0.25 ha and 40 trees & LiDAR PC & 400k pts-m2 /\newline 20k pts-m2 & No & IS & 1 & Guyana \newline Australia & Unknown \\

\rowcolor{lightgray}
AgRob V18 \cite{reis_forest_2020} & 2020 & 2019 & Unknown & LiDAR PC \newline Stereo RGB \newline Thermal \newline IMU & err +-3cm \newline N/A \newline N/A \newline N/A & No & Reg. & N/A & Portugal & Unknown \\

Montmorency dataset	\cite{tremblay_automatic_2020} & 2020 & Unknown & 1.4 ha \newline 1453 trees & LiDAR PC \newline RGB \newline DBH \newline IMU & Unknown \newline N/A \newline \newline 3.5cm \newline N/A & No & OD \newline IS \newline Reg. & 18 & Canada & Unknown \\

\rowcolor{lightgray}
ForTrunkDetV2 \cite{da_silva_visible_2021, da_silva_unimodal_2021, da_silva_edge_2022} & 2022 & 2021 \newline 2022 & 5325 images & RGB \newline Thermal & N/A \newline N/A & No & OD & 1 & Portugal & CC-BY-4.0 \\

SynthTree43k and CanaTree100 \cite{grondin_tree_2022} & 2022 & 2020 \newline 2021 & 43K synth. \newline 100 real & RGB-Depth & N/A & No & OD \newline IS \newline KD & 17 & Canada & Apache 2.0 \\

\bottomrule
\end{tabular}
\footnotetext[]{{Acronyms}: \textbf{N/A}: non applicable; \textbf{Unknown}: non provided by the authors; \textbf{ha}: hectares; \textbf{PC}: point cloud; \textbf{RGB}: red-green-blue images; \textbf{DBH}: diameter at breast height;  \textbf{IMU}: inertial measurement unit; \textbf{IS}: instance segmentation; \textbf{Reg.}: regression; \textbf{OD}: object detection; \textbf{KD}: key-point detection. Note that the dataset size measured in \textbf{K} are $O(10^3)$.}
\end{fntable}}
\label{tab:ground}
\end{table*}

\subsection{Aerial recordings}
\label{sec:review_aerial}

Aerial datasets consist of recordings of sensors mounted on unoccupied (drones) or occupied aircrafts flying above the tree canopy, offering a broader perspective of the forest without the hindrance of obstacles impeding the automatic recording process.
The diversity in aerial datasets has increased in the past few years since they are used for diverse applications such as vegetation segmentation, disease detection, fire detection and numerous others \cite{guimaraes_forestry_2020}. This is in part also boosted as governmental organizations are increasingly making the imagery of repeated official aerial campaigns openly available (\textit{e.g.} for entire countries). Furthermore, the decreasing costs of UAVs and the miniaturization of high-quality sensors have served as strong incentives for their adoption within the community.

Aerial-based recordings are reviewed in Table \ref{tab:aerial}. The dataset size is expressed in kilometer squared ($\text{km}^2$), or in hectares (ha) if the studied area is small. It is also quantified by the number of samples or number of trees if applicable.

Multiple sensors can be carried by UAVs,
including RGB and thermal cameras, multispectral sensors, hyperspectral sensors,
and LiDAR, which collectively contribute to a captivating array of recorded data, offering diverse perspectives and insights.
Cameras mounted on UAVs facilitate the acquisition of overlapping images with a spatial resolution of a few mm to cm.
Such high-resolution image datasets can be applied in concert with photogrammetric workflows, that enable a triangulation of common features found in overlapping images, enabling to precisely reconstruct camera parameters and orientations in hundreds of images automatically. Such workflows enable to reconstruct digital surface models and reprojections of the imagery to generate geocoded image mosaics with orthographic projection \cite{guimaraes_forestry_2020, diez_deep_2021}. 
Most of the recently publicly released aerial datasets contain RGB images generated by photogrammetry since they are relatively simple and cheap to collect \cite{morales_automatic_2018, kattenborn_convolutional_2019, kentsch_computer_2020, schiefer_mapping_2020, nguyen_individual_2021, kattenborn_convolutional_2020, galuszynski_automated_2022, reiersen_reforestree_2022}. But the original RGB point cloud carrying the height information used to generate the DSM is generally not provided with some exceptions \cite{brieger_advances_2019, van_geffen_sidroforest_2022}. This is unfortunate because there would be opportunities for new multi-modal models to leverage both the RGB and point cloud modalities to improve model performance.

An alternative method for studying the topography of both the ground and canopies, depending on the structure of the forest, involves the utilization of airborne LiDAR acquisitions \cite{ferraz_carbon_2018, kalinicheva_multi-layer_2022}. 
In contrast to terrestrial LiDAR, these measurements commonly have lower point densities, but cover large areas. 
Airborne LiDAR sensors are operated with IMU sensors which enables geo-referenced flights transects across large spatial extents. 
These sensors typically can record multiple returns per LiDAR pulse so that acquisitions can resemble the vertical structure of forest stands, including multiple overlapping tree layers, the understory and even the ground topography \cite{kalinicheva_multi-layer_2022}.
The spatial resolution of airborne LiDAR products is estimated by the averaged number of points per square meter.

Multispectral and hyperspectral sensors are passive, capturing reflected or emitted photons \cite{mavrovic_reviews_2023} from the sun across wavelength bands that extend beyond the visible spectrum, allowing for comprehensive recording of electromagnetic radiation throughout the near up to the shortwave infrared region.
They are especially valuable in assessing the composition of forest canopies, enabling the differentiation of species or retrieving biochemical and structural properties based on the spectral characteristics  across spectral bands \cite{fassnacht_review_2016, cherif_spectra_2023}.
A trade-off is usually required between acquiring information with a high spectral and a low spatial resolution \cite{paz-kagan_multiscale_2017}, or with a low spectral and a high spatial resolution \cite{garioud_flair_2022}, given that the radiation reflected by plant canopies does not suffice the acquisition at high spectral and high spatial resolution simultaneously.

Forest monitoring can be explored in different ways using aerial datasets relying on the sensors employed and the annotations provided alongside the data.
For instance, semantic segmentation is a prevalent method employed to classify forest canopies into tree species \cite{morales_automatic_2018, kattenborn_convolutional_2019, kentsch_computer_2020, schiefer_mapping_2020, kattenborn_convolutional_2020, galuszynski_automated_2022}. Depending on the canopy structural complexity and data quality, the classification might combined with a delineation of individual tree crowns using instance segmentation approaches. Thereby, instance segmentation captures the intricate shapes of tree crowns, unlike object detection, which typically predicts rectangular bounding boxes or centroids for individual objects \cite{reiersen_reforestree_2022}.
Some of the reviewed datasets include a DSM, which can be utilized with tree localization to estimate canopy height. 
This application using deep learning algorithms and aerial data is an actual active field of research \cite{yue_treeunet_2019, moradi_potential_2022, reiersen_reforestree_2022, wagner_sub-meter_2023}.

Due to the high spatial resolution, datasets of aerial recordings enable a granular understanding of forests at the individual tree level.
There are still many open challenges which could be explored at the tree level such as segmenting individual tree crowns in dense forests, classify them between a wide range of species or adapt algorithms from a forest to another (see Sec. \ref{sec:forest_topics_and_challenges}). 
Nevertheless, the scale of aerial datasets, especially for drones, is constrained by limited battery life and recording capacities, making it challenging to regularly assess and thus monitor large forest areas. 
Consequently, the next section will explore satellite datasets, which are better suited for capturing a broader scope of forest landscapes at high frequencies.

\begin{table*}[ht]
\fontsize{6.5pt}{7.5pt}\selectfont 
\renewcommand{\arraystretch}{1.5} 
\setlength\tabcolsep{5pt} 
\caption{Review of open access aerial forest datasets}
{\begin{fntable}
\centering
\begin{tabular}{p{1.5cm} | p{0.4cm} | p{1cm} | p{1.3cm} | p{1.2cm} : p{1.1cm} | p{0.4cm} | p{0.7cm} | p{0.8cm} | p{1cm} | p{1.1cm}}
\toprule
Dataset  & Publi. year & Recording year & Dataset size & Data  & Spatial resolution & Time series & Potential task(s) & \#Classes & Location  & License  \\
\midrule

Dorot and Negba land use \cite{paz-kagan_multiscale_2017} & 2017 & Unknown & 22.9 km2 & Hyperspectral & 1m & No & Classif. \newline Seg. & 23 & Israel & Unknown \\

\rowcolor{lightgray}
Kalimantan Lidar \cite{ferraz_carbon_2018} & 2018 & 2014 & 1.1K km2 & Lidar PC \newline CHM \newline DTM \newline DSM & 4-10 pts-m2 \newline N/A \newline N/A \newline N/A & No & Reg. & N/A & Indonesia & \href{https://www.earthdata.nasa.gov/learn/use-data/data-use-policy?}{Specific} \\

MauFlex	\cite{morales_automatic_2018} & 2018 & 2015-2018 & 25K samples & RGB & 1.4-2.5cm & Yes & Seg. & 1 & Peru & Unknown \\

\rowcolor{lightgray}
Mueller Glacier	\cite{kattenborn_convolutional_2019}  & 2019 & 2017 & 15.75 ha & RGB & 5cm & No & Seg. & 2 & New Zealand & CC-BY-4.0 \\

Cactus Aerial Photos \cite{lopez-jimenez_columnar_2019}	 & 2019 & Unknown & 21K samples & RGB & Unknown & No & Classif. & 2 & Mexico & GPL 2 \\

\rowcolor{lightgray}
Woody invasive species \cite{kattenborn_uav_2019} & 2019 & 2016/2017 & 151.7 ha & RGB \newline Hyperspectral & 3cm \newline 10cm & No & Seg. & 3 & Chile & CC-BY-4.0 \\

YURF and Shonai Coastal Forest \cite{kentsch_computer_2020} & 2020 & 2018 \newline 2019 & 2800 samples & RGB & 2.79cm to 4.48cm & Yes & MC \newline Seg. & 9 & Japan & CC-BY-4.0 \\

\rowcolor{lightgray}
TreeSeg	\cite{schiefer_mapping_2020} & 2020 & 2017-2019 & 51 ha & RGB & $\leq 2$cm & No & Seg. & 14 & Germany & CC-BY-4.0 \\

Zao Mountain \cite{nguyen_individual_2021} & 2021 & 2019 & 18 ha \newline 5354 trees & RGB & 1.5 to 2.1cm & No & OL \newline Classif. \newline Seg. & 3 & Japan & CC-BY-4.0 \\

\rowcolor{lightgray}
New Zealand primary forest \cite{kattenborn_convolutional_2020}	 & 2022 & 2017 & 4.3 ha & RGB \newline DBH \newline DSM & 3cm \newline N/A \newline N/A & No & Seg. \newline Reg. & 2 & New Zealand	 & CC-BY-4.0 \\

Portulacaria afra canopies \cite{galuszynski_automated_2022} & 2022 & 2020 \newline 2021 & 75 ha & RGB & 1cm & No & Seg. & 1 & South Africa & CC-BY-4.0 \\

\rowcolor{lightgray}
FLAIR\#1 \cite{garioud_flair_2022} & 2022 & 2019-2021 & 810 km2 \newline 77K samples & Multispectral \newline Panchromatic \newline Elevation \newline spat. DTM \newline vert. DTM & 0.2m \newline 0.2m \newline 0.4m \newline 1m \newline 0.3-7m & Yes & Seg. & 19 & France & Open Licence 2.0 \\

WildForest3D \cite{kalinicheva_multi-layer_2022} & 2022 & Unknown & 4.7 ha \newline 2000 trees & Lidar PC & 60 pts-m2 & No & Seg. & 6 & France & Unknown \\

\bottomrule

\end{tabular}
\footnotetext[]{{Acronyms}: \textbf{N/A}: non applicable; \textbf{Unknown}: non provided by the authors; \textbf{ha}: hectares; \textbf{PC}: point cloud; \textbf{CHM}: canopy height model; \textbf{DTM}: digital terrain model (spatial or vertical); \textbf{DSM}: digital surface model; \textbf{RGB}: red-green-blue images; \textbf{DBH}: diameter at breast height; \textbf{Classif.}: classification; \textbf{Seg.}: semantic segmentation; \textbf{Reg.}: regression; \textbf{MC}: multi-classification; \textbf{OD}: object detection. Note that the dataset size measured in \textbf{K} are $O(10^3)$.}
\end{fntable}}
\label{tab:aerial}
\end{table*}

\subsection{Satellite recordings}
\label{sec:review_satellite}


Satellite imagery has been consistently recorded across the globe for many years, enabling extensive research in the field of temporal remote sensing.
This abundance of data have opened up research in machine learning applied to earth observation, in particular deep learning approaches \cite{campsvalls_deep_2021}, in the past few years.
The datasets generated by diverse satellite missions encompass a wide range of resolutions and employ various sensors, enabling studies of diverse phenomena over both space and time \cite{swain_spatio-temporal_2023}.

The Landsat missions\footnote{\url{https://www.usgs.gov/landsat-missions/landsat-satellite-missions}}, a collaborative endeavor started in the seventies involving the U.S. Geological Survey (USGS), U.S. Department of the Interior (DOI), National Aeronautics and Space Administration (NASA), and the U.S. Department of Agriculture (USDA), represent the earliest and pioneering attempt to utilize multispectral cameras for Earth observation \cite{wulder_fifty_2022}.
Landsat missions 4 and 5 capture images with between 4 and 7 spectral bands, offering spatial resolutions ranging from 30 to 120 meters. 
The more recent Landsat missions, namely Landsat 7 and Landsat 8, are record images with 8 and 9 spectral bands respectively. These missions provide spatial resolutions ranging from 15 to 60 meters for Landsat 7 and 15 to 30 meters for Landsat 8.
All of the Landsat missions have a 16-day repeat cycle.
Most of the reviewed datasets used 30 meters resolution spectral bands to ensure a consistency between the bands used for their final application \cite{robinson_large_2019, potapov_annual_2019, irvin_forestnet_2020, de_almeida_pereira_active_2021, feng_m_arctic-boreal_2022, potapov_global_2022, lee_multiearth_2022}.

The Sentinel missions\footnote{\url{https://sentinel.esa.int/web/sentinel/missions}}, managed by the European Space Agency (ESA), have been designed to comprehensively monitor the Earth's various domains, encompassing air, land, ocean, and atmospheric measurements. These missions employ multiple sensors, enabling a wide range of Earth observation capabilities.
Sentinel-1 includes a SAR generating electromagnetic waves with wavelengths not impacted by clouds. The reviewed datasets provide or use Level-1 Ground Range Detected (GRD) products at a $10 \times 10$ meters resolution \cite{schmitt_sen12ms_2019, sumbul_bigearthnet-mm_2021, lee_multiearth_2022}.
Since two satellites (Sentinel-1A and Sentinel-1B) are recording data on the same orbit, the mission has a 6 days exact repeat cycle with less than a day of revisit frequency at high latitudes.

Sentinel-2 utilizes multispectral sensors to scan photon reflectance across multiple spectral bands.
The spatial resolution of the recorded data depends on the spectral bands: four bands at 10 meters, six bands at 20 meters and three bands at 60 meters. Released datasets either kept 10m resolution bands \cite{schmitt_sen12ms_2019, bastani_satlas_2022, lee_multiearth_2022} or all bands \cite{sumbul_bigearthnet-mm_2021}.
The revisit frequency of the combined constellation of Sentinel-2A and B is 5 days on most of the globe.

Data from the Landsat and Sentinel missions are the most commonly provided in the reviewed datasets, but other interesting satellite sources are also explored. For instance, the Moderate Resolution Imaging Spectroradiometer (MODIS)\footnote{\url{https://modis.gsfc.nasa.gov}} instrument, introduced by NASA and integrated into the Terra and Aqua missions, generates data that are also used for large-scale forest monitoring purposes.
The MODIS instrument offers recordings from 36 spectral bands, each defined for diverse observations, including atmospheric gases, ocean components, land boundaries, and properties \cite{schmitt_sen12ms_2019, levin_unveiling_2021}.
%
Another example is the Visible Infrared Imaging Radiometer Suite (VIIRS) instrument\footnote{\url{https://www.earthdata.nasa.gov/learn/find-data/near-real-time/viirs}}, part of the NOAA-20 missions, which also have generated data contained in a forest dataset for land and atmospheric observations \cite{levin_unveiling_2021}.
%
It should be noted that researchers have used recordings from PlanetLabs\footnote{\url{https://www.planet.com/}}, PlanetScope\footnote{\url{https://earth.esa.int/eogateway/missions/planetscope}} or Maxar\footnote{\url{https://www.maxar.com/}} missions (\textit{e.g.} GeoEye, WorldView, or QuickBird), which provide multispectral data with sub-meter spatial resolution \cite{brandt_unexpectedly_2020}. However, these datasets are not publicly accessible due to the associated licensing restrictions.


The datasets that have been reviewed encompass satellite data obtained from various missions and products, originating from different locations, and exhibiting diverse spatial and temporal resolutions.
The details of datasets published before 2020 and included, or after 2020, are provided respectively in Table \ref{tab:satellite1} and Table \ref{tab:satellite2}.
The dataset size is expressed in kilometer squared ($\text{km}^2$), or in hectares (ha) if the studied area is small. It is also quantified by the number of samples, trees or events if applicable.

Satellite datasets are frequently used for classification, multi-classification or segmentation of satellite tiles, including LULC and tree species distribution.
Other tasks include regression applications for forest cover estimation \cite{bastani_satlas_2022, feng_m_arctic-boreal_2022}, canopy height \cite{forkuor_above-ground_2020, lang_high-resolution_2022}, or live fuel moisture content estimation \cite{rao_sar-enhanced_2020}.
An intriguing application involves utilizing satellite time series data to evaluate change detection of forest covers at a large scale \cite{wang_extensive_2020}. This approach enables the estimation of deforestation, afforestation, and reforestation activities \cite{potapov_global_2022}.

Satellite recordings play a crucial role in Earth observation on a large scale, as they are manually or automatically processed to estimate global maps of forest cover, among other applications. Additionally, world maps depicting above-ground biomass, land use, and land cover have been estimated and made publicly available. In the following section, datasets containing maps at the country or global level will be reviewed.

\begin{table*}[ht]
\fontsize{6.5pt}{7.5pt}\selectfont 
\renewcommand{\arraystretch}{1.5} 
\setlength\tabcolsep{5pt} 
\caption{Review of open access satellite forest datasets before 2020 (included)}
{\begin{fntable}
\centering
\begin{tabular}{p{1.5cm} | p{0.4cm} | p{1cm} | p{1.3cm} | p{1.2cm} : p{1cm} | p{0.4cm} | p{0.7cm} | p{0.8cm} | p{1cm} | p{1.1cm}}
\toprule
Dataset  & Publi. year & Recording year & Dataset size & Data  & Spatial resolution & Time series & Potential Task(s) & \#Classes & Location  & License  \\
\midrule

GlobCover 2009	\cite{arino_olivier_global_2010} & 2010 & 2009 & 2485 samples \newline ($5^{\circ} \times 5^{\circ}$) & Multispectral & 300m & No & Classif. & 22 & Worldwide & \href{http://due.esrin.esa.int/page_globcover.php}{Specific} \\

\rowcolor{lightgray}
Spatial Database of Planted Trees \cite{harris_spatial_2019} & 2019 & 2015 & 223M ha & Multispectral & $\leq$ 30m & No & Classif. & 4 & Worldwide & CC-BY-4.0 \\

Lower Mekong \cite{potapov_annual_2019} & 2019 & 2000-2017 & 112M ha \newline 56K samples & Multispectral & 30m & Yes & Classif. \newline Reg. & 2 & Myanmar \newline Laos \newline Thailand \newline Cambodia \newline Vietnam & Unknown \\

\rowcolor{lightgray}
Chesapeake Land Cover \cite{robinson_large_2019} & 2019 & 2011-2017 & 160km2 & Multispectral & 1m \& 30m & Yes & Seg. & 4 \& 16 & USA & \href{https://www.usgs.gov/information-policies-and-instructions/copyrights-and-credits}{Specific} \newline \href{https://opendatacommons.org/licenses/odbl/}{ODbL} \\

SEN12MS	\cite{schmitt_sen12ms_2019} & 2019 & 2016-2017 & 542K samples & SAR \newline Multispectral \newline LULC maps & 10m \newline 10m \newline 10m & No & Classif. & 33 & Worldwide & CC-BY-4.0 \\

\rowcolor{lightgray}
Non-forest trees \cite{brandt_unexpectedly_2020}  & 2020 & 2005-2018 & 50K samples \newline 90K trees & Multispectral & 0.5m & No & Seg. & 1 & W. Sahara \newline Mauritania \newline Senegal \newline Gambia \newline Guinea-Bissau \newline Mali & \href{https://www.earthdata.nasa.gov/learn/use-data/data-use-policy?}{Specific} \\

ForestNet \cite{irvin_forestnet_2020} & 2020 & 2012-2020 & 76K samples \newline 2.8K events & Multispectral \newline Topography \newline Climate \newline Soil \newline Accessibility \newline Proximity & 15m/30m \newline 30m \newline 56km \newline N/A \newline N/A \newline N/A & Yes & Seg. & 4 & Indonesia & CC-BY-4.0 \\

\rowcolor{lightgray}
Live Fuel Moisture Content \cite{rao_sar-enhanced_2020} & 2020 & 2015-2019 & 164M km2 & LMFC \newline Variables \newline Location \newline Topology \newline CHM \newline LULC maps & 250m \newline 250m \newline 250m \newline 250m \newline 250m \newline 250m & Yes & Classif. \newline Reg. & 6 & USA & CC-BY-NC-ND-4.0 \\

\bottomrule

\end{tabular}
\footnotetext[]{{Acronyms}: \textbf{N/A}: non applicable; \textbf{ha}: hectares; \textbf{SAR}: synthetic-aperture RADAR; \textbf{LULC}: land use and/or land cover; \textbf{CHM}: canopy height model; \textbf{LMFC}: live fuel moisture content; \textbf{Classif.}: classification; \textbf{Seg.}: semantic segmentation; \textbf{Reg.}: regression.
Note that the dataset size measured in \textbf{K} are $O(10^3)$ and in \textbf{M} are $O(10^6)$.
}
\end{fntable}}
\label{tab:satellite1}
\end{table*}

\begin{table*}[ht]
\fontsize{6.5pt}{7.5pt}\selectfont 
\renewcommand{\arraystretch}{1.5} 
\setlength\tabcolsep{5pt} 
\caption{Review of satellite recording datasets after 2021 (included)}
{\begin{fntable}
\centering
\begin{tabular}{p{1.5cm} | p{0.4cm} | p{1cm} | p{1.3cm} | p{1.2cm} : p{1cm} | p{0.4cm} | p{0.7cm} | p{0.8cm} | p{1cm} | p{1.1cm}}
\toprule
Dataset  & Publi. year & Recording year & Dataset size & Data  & Spatial resolution & Time series & Potential Task(s) & \#Classes & Location  & License  \\
\midrule

Australian Black Summer	\cite{levin_unveiling_2021} & 2021 & 2001-2021 & 265K km2 \newline 391 fires & Locations \newline Variables & 500m \newline 500m & No & Seg. & 3 & Australia	 & CC-BY-4.0 \\

\rowcolor{lightgray}
Active fire detection \cite{de_almeida_pereira_active_2021} & 2021 & 2020 & 150K samples & Multispectral & 30m & No & Seg. & 1 & Worldwide & Unknown \\

BigEarthNet-MM	\cite{sumbul_bigearthnet-mm_2021} & 2021 & 2017 \newline 2018 & 590K samples ($\times 2$) & SAR \newline Multispectral	& 10m \newline 10m to 60m & No & MC & 19 & 10 countries (Europe)$\star$& \href{https://bigearth.net/downloads/documents/License.pdf}{Specific} \\

\rowcolor{lightgray}
Satlas \cite{bastani_satlas_2022} & 2022 & 2016-2021 & 85M km2 \newline 5M megapixels & Multispectral & 10m / 1m & Yes & Seg. \newline Classif. \newline Reg. & 137 & Worldwide & Apache License 2.0 \\

GLAD Global Land Cover and Land Use change	\cite{potapov_global_2022} & 2022 & 2000-2020 & 504 samples \newline ($10^{\circ} \times 10^{\circ}$) & Multispectral \newline LULC maps & 30m \newline $1 \times 1^{\circ}$ & Yes & Classif. \newline Reg. \newline CD & 5 & Worldwide & CC-BY-4.0 \\

\rowcolor{lightgray}
ABoVE \cite{feng_m_arctic-boreal_2022} & 2022 & 1984-2020 & 224K samples & Multispectral & 30m & Yes & Reg. & N/A & Worldwide & \href{https://www.earthdata.nasa.gov/learn/use-data/data-use-policy?}{Specific} \\

MultiEarth 2022 Deforestation Challenge	\cite{lee_multiearth_2022} & 2022 & 2014-2021 \newline 2018-2021 \newline 1984-2012 \newline 2013-2021 & 12M samples & SAR \newline Multispectral & 10m \newline 10m / 30m & Yes & Seg. & 1 & Brazil & CC-BY-4.0 \\

\rowcolor{lightgray}
EarthNet2021x \cite{robin_learning_2022} & 2022 & 2016-2021 & Unknown & NDVI map \newline Multispectral \newline DEM \newline LULC map \newline Geomorpho. \newline Meterology \newline Land surface soil moisture & 30m \newline 30m \newline 30m \newline 30m \newline 90m \newline Unknown \newline Unknown & Yes & Reg. & N/A & Africa	 & CC-BY-NC-S-4.0 \\

\bottomrule

\end{tabular}
\footnotetext[]{{$\star$}: The list of countries is detailed in the OpenForest catalogue.}
\footnotetext[]{{Acronyms}: \textbf{N/A}: non applicable; \textbf{Unknown}: non provided by the authors; \textbf{ha}: hectares; \textbf{SAR}: synthetic-aperture RADAR; \textbf{LULC}: land use and/or land cover; \textbf{NDVI}: normalized difference vegetation index; \textbf{Classif.}: classification; \textbf{Seg.}: semantic segmentation; \textbf{Reg.}: regression; \textbf{MC}: multi-classification; \textbf{CD}: change detection.
Note that the dataset size measured in \textbf{K} are $O(10^3)$ and in \textbf{M} are $O(10^6)$.}
\end{fntable}}
\label{tab:satellite2}
\end{table*}

\subsection{Country or world maps}
\label{sec:review_maps}


Earth observation applications have been resumed into maps at the country, continent, or global level. They are estimated using machine learning algorithms that incorporate manual expert features derived from satellite data such as statistics of the data distribution or vegetation indexes. These features encompass various aspects, ranging from multispectral information \cite{friedl_modis_2010, pflugmacher_mapping_2019} to climatic and elevation data \cite{chaves_mapping_2020}.
The majority of the released maps have been estimated using machine learning algorithms trained on satellite data, as these algorithms demonstrate excellent scalability for predicting at a large scale and low resolution.
The results obtained from these algorithms have been validated using field inventories. However, it is important to note that the field inventories themselves have not been included in the reviewed datasets discussed in this section\footnote{Open access datasets releasing both maps and inventories are reviewed in Section \ref{sec:review_mixed}, Table \ref{tab:mixed1}.}. 
Nonetheless, the reviewed map datasets are notable for their global coverage, which adds to their significance.
Despite containing inherent uncertainties in their estimations, these maps have the potential to offer valuable meta-knowledge to downstream applications in the realm of forest monitoring.

Large scale maps datasets before 2019 (included) are reviewed in Table \ref{tab:maps1} while maps datasets released after 2020 (included) are reviewed in Table \ref{tab:maps2}.
The dataset size is expressed in kilometer squared ($\text{km}^2$), or in hectares (ha) if the studied area is small. It is also quantified by the number of samples or points if applicable.

A significant portion of map datasets focuses on providing information about LULC (see Sec. \ref{sec:review} for the proposed definition), which plays a crucial role in distinguishing different types of forests at a large scale \cite{bartholome_glc2000_2005, friedl_modis_2010, griffiths_forest_2014, pflugmacher_mapping_2019, thonfeld_long-term_2020, bonannella_forest_2022}.
Within in LULC maps, the reviewed works have also estimated the extend of forest cover, including time series data.
These time series are particularly valuable for quantifying forest loss, \textit{i.e.} deforestation detection, as well as forest gain, \textit{i.e.} afforestation, reforestation monitoring or deadwood maps \cite{hansen_high-resolution_2013, curtis_classifying_2018, verhegghen_mapping_2022, bunting_global_2022, schiefer_uav-based_2023}.
Another category of maps is specifically designed to differentiate between different plant functional types, particularly distinguishing between broad-leaved and needle-leaf forests, as well as identifying summer-green and evergreen forests across tropical, boreal, and temperate regions \cite{ottle_use_2013}.


As mentioned in Section \ref{sec:forest_challenges}, accurate estimation of above-ground biomass is crucial for a comprehensive quantification of the carbon stocks that forests worldwide hold.
%
World maps of above ground biomass have been estimated at different resolutions \cite{patterson_statistical_2019, tang_high-resolution_2021, ma_high-resolution_2021, santoro_global_2021}.
%
Quantifying the uncertainty of above-ground biomass maps is also important as it depends on multiple factors, including tree species, tree height, canopy size, and reference data distribution \cite{patterson_statistical_2019, santoro_global_2021, ploton2020spatial}.
Canopy height maps have also been quantified in sparse boreal forests
\cite{bartsch_land_2016, bartsch_feasibility_2020}.
These canopy height maps, both at the country \cite{tolan_sub-meter_2023} or world level \cite{lang_global_2022, lang_high-resolution_2022}, have been estimated using LiDAR sensors as a ground truth.
%
Accurate estimation of canopy height is crucial for evaluating above-ground biomass, which is why a few studies have utilized data from the Global Ecosystem Dynamics Investigation (GEDI) mission\footnote{\url{https://gedi.umd.edu/}} \cite{patterson_statistical_2019, tang_high-resolution_2021, ma_high-resolution_2021, lang_global_2022, tolan_sub-meter_2023}. 
GEDI records LiDAR data from the International Space Station, allowing for the estimation of a DSM that serves as a valuable reference for canopy height estimation.
The overall biomass estimation of forests also includes below ground biomass \cite{chen_maps_2023} and soil carbon stock \cite{dionizio_carbon_2020} which have been quantified using SAR satellite data penetrating dense canopies.

Although open access map datasets are subject to the limitations and uncertainties inherent in the estimation methods employed by the authors, they remain a valuable source of data for obtaining a broad-scale understanding of forests or integrating meta-knowledge into future analyses. These datasets could be helpful for conducting further research and expanding our knowledge of forest ecosystems.

In the preceding sections, the reviewed datasets were presented with a focus on different scales. However, in the forthcoming section, datasets that offer a combination of data at various scales will be discussed in detail.

\begin{table*}[ht]
\fontsize{6.5pt}{7.5pt}\selectfont 
\renewcommand{\arraystretch}{1.5} 
\setlength\tabcolsep{5pt} 
\caption{Review of open access map forest datasets before 2019 (included)}
{\begin{fntable}
\centering
\begin{tabular}{p{1.5cm} | p{0.4cm} | p{1cm} | p{1.3cm} | p{1.2cm} : p{1cm} | p{0.4cm} | p{0.7cm} | p{0.8cm} | p{1cm} | p{1.1cm}}
\toprule
Dataset  & Publi. year & Recording year & Dataset size & Data  & Spatial resolution & Time series & Potential task(s) & \#Classes & Location  & License  \\
\midrule

GCL2000	\cite{bartholome_glc2000_2005} & 2005 & 1999-2000 & 148M km2 & LULC maps & $1/112^{\circ}$ & Yes & Classif. \newline Seg. & 22 & Worldwide & CC-BY-4.0 \\

\rowcolor{lightgray}
MODIS MCD12Q1 \cite{friedl_modis_2010} & 2010 & 2001-2023 & Unknown & LULC maps	& 500m & Yes & Classif. \newline Seg. & 12 & Worldwide & \href{https://lpdaac.usgs.gov/data/data-citation-and-policies/}{Specific} \\

Carpathian ecoregion \cite{griffiths_forest_2014} & 2010 & from 1985 to 2010 $\blacklozenge$ & 390K km2	& Disturbance \newline LULC maps & 900m & Yes & Classif. \newline Seg. & 10 & Czechia \newline Austria \newline Poland \newline Hungary \newline Ukraine \newline Romania \newline Slovakia & CC-BY-3.0 \\

\rowcolor{lightgray}
Forest Cover Change	\cite{hansen_high-resolution_2013} & 2013 & 2000-2021 & 504 samples \newline  ($10^{\circ} \times 10^{\circ}$) & LULC maps & 30m & Yes & Reg. & N/A & Worldwide & CC-BY-4.0 \\

Siberian plant functional type \cite{ottle_use_2013} & 2013 & 2000-2010 & 5M km2 & PFT maps & 1km & No & Classif. \newline Seg. & 16 & Russia & CC-BY-3.0 \\

\rowcolor{lightgray}
Intact Forest Landscape	\cite{potapov_last_2017} & 2017 & 2000 \newline 2013 \newline 2016 \newline 2020 & 58M km2 & IFL maps & 10km & Yes & Classif. \newline Seg. & 4 & Worldwide & CC-BY-4.0 \\

Global Forest Loss	\cite{curtis_classifying_2018} & 2018 & 2000-2015 & 65K km2 & LULC maps & 10km & Yes & Classif. \newline Seg. & 5 & Worldwide & Unknown \\

\rowcolor{lightgray}
Hybrid estimators of AGB \cite{patterson_statistical_2019} & 2019 & 2019-2021 & 10 maps & AGB maps & 1km & Yes & Reg. & N/A & Worldwide & \href{https://www.earthdata.nasa.gov/learn/use-data/data-use-policy?}{Specific} \\

Pan-Europe land cover \cite{pflugmacher_mapping_2019} & 2019 & 2014-2016 & Unknown & LULC maps & 30m & No & Classif. \newline Seg. & 12 & 28 countries (Europe)$\star$ & CC-BY-SA-4.0 \\

\bottomrule

\end{tabular}
\footnotetext[]{{$\blacklozenge$}: The list of recording years is detailed in the OpenForest catalogue.}
\footnotetext[]{{$\star$}: The list of countries is detailed in the OpenForest catalogue.}
\footnotetext[]{{Acronyms}: \textbf{N/A}: non applicable; \textbf{Unknown}: non provided by the authors; \textbf{LULC}: land use and/or land cover; \textbf{PFT}: plant functional type; \textbf{IFL}: intact forest landscape; \textbf{AGB}: above ground biomass; \textbf{Classif.}: classification; \textbf{Seg.}: semantic segmentation; \textbf{Reg.}: regression.
Note that the dataset size measured in \textbf{K} are $O(10^3)$ and in \textbf{M} are $O(10^6)$.}
\end{fntable}}
\label{tab:maps1}
\end{table*}

\begin{table*}[ht]
\fontsize{6.5pt}{7.5pt}\selectfont 
\renewcommand{\arraystretch}{1.5} 
\setlength\tabcolsep{5pt} 
\caption{Review of open access map forest datasets after 2020 (included)}
{\begin{fntable}
\centering
\begin{tabular}{p{1.5cm} | p{0.4cm} | p{1cm} | p{1.3cm} | p{1.2cm} : p{1cm} | p{0.4cm} | p{0.7cm} | p{0.8cm} | p{1cm} | p{1.1cm}}
\toprule
Dataset  & Publi. year & Recording year & Dataset size & Data  & Spatial resolution & Time series & Potential task(s) & \#Classes & Location  & License  \\
\midrule

Peruvian Amazonia floristic patterns \cite{chaves_mapping_2020} & 2020 & 2013-2018 & 790K km2 & Floristic maps & 450m & No & MC & 10 & Peru & CC-BY-4.0 \\

\rowcolor{lightgray}
Florida mangrove resilience \cite{lagomasino_storm_2020} & 2020 & 2017 & 1.3 km2 & Mangrove resilience maps  & 30m & No & Classif. \newline Seg. & 3 & USA & CC-BY-4.0 \\

Cerrado biome \cite{dionizio_carbon_2020} & 2020 & 1990-2018 & 131K km2 & AGB maps \newline BGB maps \newline SCS maps & 30m \newline 30m \newline 30m & Yes & Reg. & N/A & Brazil & CC-BY-4.0 \\

\rowcolor{lightgray}
Arctic trees height	\cite{bartsch_land_2016, bartsch_feasibility_2020} & 2020 & from 2005 to 2018 $\blacklozenge$ & Unknown & CH maps & 20m & Yes & Reg. & N/A & Russia \newline  USA \newline Canada & CC-BY-4.0 \\

Kilombero Valley land cover	\cite{thonfeld_long-term_2020} & 2020 & 1974 \newline 1994 \newline 2004 \newline 2014 \newline 2015 & 40.2km2 \newline 32.7K points & LULC maps \newline LULC points & 30m \newline 1cm & Yes & Classif. \newline Seg. & 11 & Tanzania & CC-BY-4.0 \\

\rowcolor{lightgray}
Pleroma trees \cite{wagner_flowering_2021} & 2021 & 2016-2020 & 2M km2 & LULC maps & 1.28km2 & Yes & Classif. & 1 & Brazil & CC-BY-4.0 \\

Global Forest AGB 2010 \cite{santoro_global_2021} & 2021 & 2010 & 58 samples \newline ($40^{\circ} \times 40^{\circ}$) & GSV maps \newline AGB maps & 1ha \newline 1ha & No & Reg. & N/A & Worldwide & CC-BY-4.0 \\

\rowcolor{lightgray}
New England AGB	\cite{tang_high-resolution_2021, ma_high-resolution_2021} & 2021 & 2010-2015 & 187K km2 & AGB maps \newline CH maps \newline LULC maps & 30m \newline 30m \newline 30m & No & Reg. & N/A & USA & \href{https://www.earthdata.nasa.gov/learn/use-data/data-use-policy?}{Specific} \\

Dry tropical forest cover \cite{verhegghen_mapping_2022} & 2022 & 2018 & 953.5K km2 & LULC maps & 10m & No & Classif. \newline Seg. & 9 & Tanzania & CC-BY-4.0 \\

\rowcolor{lightgray}
Global Mangrove Watch (v3.0) \cite{bunting_global_2022} & 2022 & from 1996 to 2020 $\blacklozenge$ & 152K km2 \newline 17K points  & LULC maps & 25m & Yes & CD & 4 & Worldwide & CC-BY-4.0 \\

European Trees \cite{bonannella_forest_2022} & 2022 & from 2000 to 2020 $\blacklozenge$  & 4.4M points & LULC maps & 30m & Yes & Classif. \newline Seg. & 16 & Europe & CC-BY-4.0 \\

\rowcolor{lightgray}
Global canopy height \cite{lang_high-resolution_2022, lang_global_2022} & 2022 & 2019/2020 & 600M samples & CH map \newline LULC map & 10m \newline 10m & No & Reg. & N/A & Worldwide & CC-BY-4.0 \\

AGBC and BGBC of China \cite{chen_maps_2023} & 2023 & 2002-2021 & 9.6M km2 & AGB maps \newline BGB maps & 1km \newline 1km & Yes & Reg. & N/A & China & CC-BY-4.0 \\

\rowcolor{lightgray}
High-resolution canopy height map \cite{tolan_sub-meter_2023} & 2023 & 2017-2020 & Unknown & CH maps & 0.5m & No & Reg. & N/A & USA \newline Brazil & CC-BY-4.0 \\

Deadwood cover \cite{schiefer_uav-based_2023} & 2023 & 2018 \newline 2019 \newline 2020 \newline 2021 & 727.33ha & Deadwood map & 10m & No & Classif. & 1 & Germany \newline Finland & CC-BY-NC 4.0 \\

\bottomrule

\end{tabular}
\footnotetext[]{{$\blacklozenge$}: The list of recording years is detailed in the OpenForest catalogue.}
\footnotetext[]{{Acronyms}: \textbf{N/A}: non applicable; \textbf{Unknown}: non provided by the authors; \textbf{AGB}: above ground biomass; \textbf{BGB}: below ground biomass; \textbf{SCS}: soil carbon stock; \textbf{LULC}: land use and/or land cover; \textbf{GSV}: growing stock volume; \textbf{CH}: canopy height; \textbf{Classif.}: classification; \textbf{Seg.}: semantic segmentation; \textbf{Reg.}: regression; \textbf{MC}: multi-classification; \textbf{CD}: change detection.
Note that the dataset size measured in \textbf{K} are $O(10^3)$ and in \textbf{M} are $O(10^6)$.}
\end{fntable}}
\label{tab:maps2}
\end{table*}

\subsection{Datasets mixed at different scales}
\label{sec:review_mixed}

Datasets that offer data at various scales play an important role in establishing a bridge between different modalities recorded by diverse sensors. 
By integrating information from multiple sources, these datasets facilitate a comprehensive understanding of forests and enable cross-modal analysis.
%
%
Inventories, ground-based and aerial-based datasets are available at small scale but usually come alongside precise annotations at all tree levels.
Conversely, satellite and maps datasets are available at a larger scale but often lack precise annotations due to their lower resolution. 
Integrating data from different scales can be advantageous in generalizing and extrapolating local knowledge to a larger scale, bridging the gap between detailed annotations and broader coverage \cite{kattenborn_uav_2019, schiefer_uav-based_2023}.

The size of each dataset is expressed in kilometer squared ($\text{km}^2$), or in hectares (ha) if the studied area is small. It is also quantified by the number of samples, points or trees if applicable.

Mixed datasets composed of inventories and aerial-based recordings (IA); inventories, aerial-based and satellite-based recordings (IAS); and inventories and maps (IM) are reviewed in Table \ref{tab:mixed1}.
Inventories provide an additional value to imagery recordings by providing geo-located annotations, depending on the level of precision they offer.
These inventories enhance the spatial context and accuracy of the annotations.
Combining inventories with aerial-based recordings would be highly beneficial for accurately aligning tree measurements with aerial data, especially with LiDAR \cite{weiser_individual_2022} or RGB \cite{brieger_advances_2019, van_geffen_sidroforest_2022} recordings. This integration enables improved estimation of carbon stocks at the aerial scale, among other applications.
Field measurements have also been used to validate country or world maps, they are often released together to enable reproducibility of the results.
This integration of field measurements and map data enhances the accuracy and reliability of the generated maps.
Similarly to datasets presented in Section \ref{sec:review_maps}, maps of 
forest age \cite{besnard_mapping_2021}, carbon stocks \cite{tucker_sub-continental-scale_2023}, land use and land cover \cite{koskinen_participatory_2019, bendini_combining_2020, bendini_combining_2020, shevtsova_strong_2020, european_commission_statistical_office_of_the_european_union_lucas_2021} have been estimated by machine learning algorithms while being calibrated and validated with inventories.

Mixed datasets composed of ground-based and aerial-based recordings (GA); aerial-based and satellite-based recordings (AS); aerial-based recordings and maps (AM); and satellite-based recordings and maps (SM) are reviewed in Table \ref{tab:mixed2}.
%
Aligning ground-based and aerial-based imagery recordings is valuable in integrating information from both above and below the canopies of a forest. For instance, models can be trained using ground recordings sourced from citizen science-based photographs, and then effectively transferred to aerial data \cite{soltani_transfer_2022}. 
This alignment could enable a comprehensive understanding of the forest ecosystem by bridging the gap between ground-level and aerial-level observations.
%

Mapping aerial-based and satellite-based recordings is helpful for extrapolating high-resolution information at a small scale to a lower resolution at a larger scale. 
This process allows for the transfer of detailed information captured through aerial LiDAR, for example, to validate canopy height models derived from satellite imagery \cite{marconi_data_2019, weinstein_remote_2021, lang_high-resolution_2022}. The integration of these datasets facilitates a more comprehensive and accurate representation of forest characteristics across different spatial scales.
%
Integrating SAR and multispectral satellite data with aerial imagery can potentially enhance model performances \cite{schmitt_data_2016}, particularly by leveraging the varying reflection and absorption characteristics of different tree species \cite{ahlswede_treesatai_2022}.
Aerial LiDAR metrics have also been used as validation points to estimate the above ground biomass at large scale \cite{hudak_carbon_2020}.
At a larger scale, satellite data analyzed with multi-classification algorithms have also been useful to monitor and detect forest loss \cite{turubanova_ongoing_2018}.

Open access datasets featuring modalities at different scales have been made available to enable result reproducibility and promote heterogeneity in the way of observing forests. 
These datasets incorporate various modalities aligned at different scales, which could aim to enhance the generalization capabilities of machine learning algorithms at a larger scale. 
This not only facilitates research in solving tasks at different scales depending on the modality but also fosters a comprehensive understanding of forests through multi-modal analysis.
%
To date, the publications related to the reviewed datasets have not extensively explored multi-modal (\textit{e.g.} point clouds with raster data, point observations with spatially continuous data), multi-scale, and multi-task approaches. 
However, it is our hope that the machine learning and computer vision communities will venture into forest monitoring along this path, as it holds great potential for advancing our understanding of the composition of forests worldwide. 
By embracing these comprehensive approaches, we can enhance our comprehension of forests and contribute to more effective and efficient forest management strategies.
%
The upcoming section will explore perspectives on forest datasets, shedding light on the potential challenges that researchers could prioritize and address in their work.

\begin{table*}[ht]
\fontsize{6.5pt}{7.5pt}\selectfont 
\renewcommand{\arraystretch}{1.5} 
\setlength\tabcolsep{5pt} 
\caption{Review of open access mixed forest datasets, including: inventories and aerial-based (IA); inventories, aerial-based and satellite-based (IAS); inventories and maps (IM)}
{\begin{fntable}
\centering
\begin{adjustwidth}{-0.8cm}{}
\begin{tabular}{p{1.4cm} | p{0.3cm} | p{0.4cm} | p{1cm} | p{1.4cm} | p{1.4cm} : p{1.5cm} | p{0.4cm} | p{0.7cm} | p{1cm} | p{1cm} | p{1.1cm}}
\toprule
Dataset  & Type & Publi. year & Recording year & Dataset size & Data  & Spatial resolution or precision & Time series & Potential task(s) & \#Classes & Location  & License  \\
\midrule


ReforestTree \cite{reiersen_reforestree_2022} & IA & 2022 & 2020 & 4663 trees \newline 3ha & Location \newline Height \newline DBH \newline Biomass \newline Aerial RGB & <1cm \newline N/A \newline N/A \newline N/A \newline 2cm & No & OL \newline Classif. \newline Reg. \newline OD & 2 & Ecuador & CC-BY-4.0 \\

\rowcolor{lightgray}
Individual Tree Point Clouds \cite{weiser_individual_2022} & IA & 2022 & 2019 & 58.2 ha \newline 1491 trees & Location \newline Lidar PC \newline DBH \newline Height \newline Crown diam. \newline CBH  &  10cm \newline $\bullet$ pts-m2 \newline N/A \newline N/A \newline N/A \newline N/A & No & Seg. \newline IS \newline Reg. & 22 & Germany & CC-BY-SA-4.0 \\


Canopy traits \cite{cherif_spectra_2023} & IA & 2023 & Unknown & 5573 samples & Hyperspectral & 0.2m to 5m & No & Reg. & N/A & 7 countries (world) $\star$ & Unknown \\

\rowcolor{lightgray}
SiDroForest (4 datasets) \cite{brieger_advances_2019, van_geffen_sidroforest_2022} & IAS & 2022 & 2018 & 19.3K crowns \newline 872 trees \newline 10K synth. \newline 550 samples & Location \newline Height \newline Crown diam. \newline $\dagger$ Aerial record. \newline Sat. Multispec. & 10cm \newline N/A \newline N/A \newline 3cm \newline 10m & Yes & OL \newline Classif. \newline Seg. \newline Reg. & 1 \newline 11 \newline 2 \newline 11 & Russia & CC-BY-4.0 \\

South Korea land cover \cite{seo_deriving_2014} & IM & 2014 & 2009-2011 & 64.4 km2 \newline 3.4k polygones & Polygones and polylines & 1m & No & MC \newline Seg. & 67 & South Korea	& CC-BY-NC-3.0 \\ 

\rowcolor{lightgray}
Tazmanian tree plantation \cite{koskinen_participatory_2019} & IM & 2019 & 2010 \newline 2013-2016 & 7500 points \newline 250k km2 & Location \newline LULC maps & 10cm \newline 30m & No & Classif. & 4 & Tanzania	 & CC-BY-4.0 \\

Brazilian Savanna \cite{bendini_combining_2020} & IM & 2020 & from 2008 to 2019 $\blacklozenge$ & 2828 trees \newline Unknown & Location \newline LULC maps & <1cm \newline 30m & No & Classif. & 3 & Brazil & CC-BY-4.0 \\

\rowcolor{lightgray}
Siberia land cover \cite{shevtsova_strong_2020} & IM & 2020 & from 2000 to 2017 $\blacklozenge$ & 2696 points & Location \newline LULC maps & 1cm \newline 30m & Yes & Classif. & 4 & Russia & CC-BY-4.0 \\

Pyrenean oak forest	\cite{perez-luque_land-use_2021} & IM & 2021 & 2016 & 232 km2 \newline 4347 trees & Location \newline DBH \newline Height \newline EVI maps & 1cm \newline N/A \newline N/A \newline 250m & No & Reg. & Unknown & Spain & CC-BY-4.0 \\

\rowcolor{lightgray}
ForestAgeBGI \cite{besnard_mapping_2021} & IM & 2021 & 2000-2019 & 44K trees & Field measur. \newline Climatic var. \newline LULC maps \newline AGB \newline Disturbance & 1km \newline 1km \newline 1km \newline 1km \newline 1km & No & Regress. & N/A & Worldwide & CC-BY-4.0 \\

LUCAS \cite{european_commission_statistical_office_of_the_european_union_lucas_2021} & IM & 2021 & from 2009 to 2018 $\blacklozenge$ & 1.6M points & Location \newline LULC maps & >1cm \newline 2km2 & No & Classif. & 12 \newline 114 subclasses & 27 countries (Europe)$\star$ & \href{https://esdac.jrc.ec.europa.eu/projects/LUCAS/Documents/LUCAS_SOIL_LIC_AGR_final_for_web.pdf}{Specific} \\

\rowcolor{lightgray}
Sub-Saharan carbon stocks \cite{tucker_sub-continental-scale_2023} & IM & 2023 & 2002-2020 & 9.9B trees \newline 10M km2 & $\ddagger$ Field measur. \newline LULC maps & 50cm \newline 50cm & No & Seg. \newline Reg. & 1	& 26 countries (Africa)$\star$ & \href{https://daac.ornl.gov/about/#citation_policy}{Specific} \\

\bottomrule

\end{tabular}
\end{adjustwidth}
\footnotetext[]{{$\blacklozenge$}: The list of recording years is detailed in the OpenForest catalogue.}
\footnotetext[]{{$\bullet$}: The dataset includes three LiDAR with different resolutions which are ALS: 72.5pts-m2, ULS: 1029.2pts-m2, TLS: Unknown.}
\footnotetext[]{{$\dagger$}: Aerial recordings (3cm resolution) are aerial RGB, SfM PC, RGB PC, RGN images, DEM, CHM, DSM, DTM.}
\footnotetext[]{{$\ddagger$}: Field measurements (50cm resolution) are location, crown area, wood mass, mass, root dry mass, count density, coverage density, area density.}
\footnotetext[]{{$\star$}: The list of countries is detailed in the OpenForest catalogue.}
\footnotetext[]{{Acronyms}: \textbf{IA}: inventories and aerial; \textbf{IAS}: inventories, aerial and satellite; \textbf{IM}: inventories and maps; \textbf{N/A}: non applicable; \textbf{Unknown}: non provided by the authors; \textbf{RGB}: red-green-blue; \textbf{DBH}: diameter at breast height; \textbf{PC}: point cloud; \textbf{CBH}: crown base height; \textbf{LULC}: land use and/or land cover; \textbf{EVI}: enhanced vegetation index; \textbf{AGB}: above ground biomass; \textbf{OL}: object localization; \textbf{Classif.}: classification; \textbf{Seg.}: semantic segmentation; \textbf{IS}: instance segmentation; \textbf{Reg.}: regression; \textbf{MC}: multi-classification.
Note that the dataset size measured in \textbf{K} are $O(10^3)$, in \textbf{M} are $O(10^6)$ and in \textbf{B} are $O(10^9)$.}
\end{fntable}}
\label{tab:mixed1}
\end{table*}

\begin{table*}[ht]
\fontsize{6.5pt}{7.5pt}\selectfont 
\renewcommand{\arraystretch}{1.5} 
\setlength\tabcolsep{5pt} 
\caption{Review of open access mixed forest datasets, including: ground-based and aerial-based (GA); aerial-based and satellite-based (AS); aerial-based and maps (AM); satellite-based and maps (SM)}
{\begin{fntable}
\centering
\begin{adjustwidth}{-0.7cm}{}
\begin{tabular}{p{1.2cm} | p{0.3cm} | p{0.4cm} | p{1cm} | p{1.4cm} | p{1.4cm} : p{1.5cm} | p{0.4cm} | p{0.7cm} | p{1cm} | p{1cm} | p{1.1cm}}
\toprule
Dataset  & Type & Publi. year & Recording year & Dataset size & Data  & Spatial resolution or precision & Time series & Potential task(s) & \#Classes & Location  & License  \\
\midrule

Knotweed UAV imagery \cite{soltani_transfer_2022} & GA & 2022 & 2021 & 20 km2 \newline 10K samples & Ground RGB \newline Aerial RGB  & N/A \newline 0.3cm & No & Seg. & 1 & Germany & CC-BY-4.0 \\

\rowcolor{lightgray}
IDTReeS	\cite{marconi_data_2019} & AS & 2021 & 2014 \newline 2015 & 0.344 ha & Lidar PC \newline Aerial hypersp. \newline Lidar CHM \newline Sat. RGB & 5pts-m2 \newline 1m \newline 1m \newline 10cm & No & Classif. \newline Seg. \newline Align. & 9 & USA & CC-BY-4.0 \\

NeonTree Evaluation	\cite{weinstein_benchmark_2021} & AS & 2021 & 2017 \newline 2018 \newline 2019 & 31K trees & Lidar PC \newline Aerial hypersp. \newline Sat. RGB & 5pts-m2 \newline 1m \newline 10cm & No & OD \newline Seg. & 1 & USA & CC-BY-4.0 \\

\rowcolor{lightgray}
NEON Crowns	\cite{weinstein_remote_2021} & AS & 2021 & 2018 \newline 2019 & 104M trees & Lidar PC \newline Height \newline Sat. RGB & 5pts-m2 \newline N/A \newline 10cm & No & OD  & 1 & USA & CC-BY-4.0 \\

TreeSatAI \cite{ahlswede_treesatai_2022} & AS & 2022 & 2011-2020 \newline 2015-2020 & 50K samples & Aerial RGB \newline Sat. SAR \newline Sat. Multispec. & 20cm \newline 10m \newline 10m & Yes & MC & 20 & Germany	& CC-BY-4.0 \\

\rowcolor{lightgray}
CMS AGB USA	\cite{hudak_carbon_2020} & AM & 2020 & 2000-2016 & 130K km2 \newline 97K trees & Lidar PC \newline AGB maps & 5pts-m2 \newline 30m & Yes & Reg. & N/A & USA & \href{https://www.earthdata.nasa.gov/learn/use-data/data-use-policy?}{Specific} \\

Primary Forest Loss	\cite{turubanova_ongoing_2018} & SM & 2018 & 2002-2014 & 49K samples & Sat. Multispec. \newline NDVI maps \newline LULC maps & 30m \newline 30m \newline 30m & Yes & MC & 4 & Brazil \newline Dem. Rep. Congo \newline Indonesia & CC-BY-4.0 \\

\bottomrule

\end{tabular}
\end{adjustwidth}
\footnotetext[]{{Acronyms}: \textbf{GA}: ground and aerial; \textbf{AS}: aerial and satellite; \textbf{AM}: aerial and maps; \textbf{SM}: satellite and maps; \textbf{N/A}: non applicable; \textbf{Unknown}: non provided by the authors; \textbf{RGB}: red-green-blue; \textbf{PC}: point cloud; \textbf{CHM}: canopy height model; \textbf{SAR}: synthetic-aperture RADAR; \textbf{AGB}: above ground biomass; \textbf{NDVI}: normalized difference vegetation index; \textbf{LULC}: land use and/or land cover; \textbf{Classif.}: classification; \textbf{Seg.}: semantic segmentation; \textbf{Align.}: alignment; \textbf{OD}: object detection; \textbf{Reg.}: regression; \textbf{MC}: multi-classification.
Note that the dataset size measured in \textbf{K} are $O(10^3)$ and in \textbf{M} are $O(10^6)$.}
\end{fntable}}
\label{tab:mixed2}
\end{table*}


\section[Perspectives]{Perspectives}
\label{sec:perspectives}


The enthusiasm for forest monitoring is on the rise, serving as a safeguard to protect forests and their ecological and societal significance.
Proper monitoring is essential for avoided forest conversion, supporting forest management initiatives, and ensuring successful reforestation and afforestation projects by enhancing survival rates and preventing diseases \cite{van_lierop_global_2015, martin_people_2021}. Additionally, the effects of climate change on forest dynamics \cite{fassnacht_remote_2023} imply a growing need for heightened surveillance of these ecosystems.
%
%

As a data-driven and empirical science, respectively, forest monitoring benefits from open access, diverse and large datasets, coupled with advancements in machine learning research \cite{de_lima_making_2022}.
This endeavor seeks to address existing challenges and research strategies while extensively reviewing open access forest datasets, with the ultimate goal of encouraging the research community to further investigate this field.

As evident from Section \ref{sec:forest_challenges}, forest monitoring remains an active area of research. Numerous ongoing inquiries delve into various aspects, such as tree species identification, phenology, abiotic factors, exogenous influences and many more.
Machine learning already greatly advanced our capabilities to monitor forests through novel analytical tools and capacities. 
This involves sensing past, current and dynamic forest states through predictive modelling. 
Such models and information, build to be explainable by design, will greatly advance our understanding of forests, including insights into how diverse environmental and anthropogenic drivers impact forest dynamics, as well as the operational mechanisms of forest ecosystems.
%
A related and cardinal interest lies in the projection of future forest dynamics to guide decision makers, to improve management and anticipate consequences \cite{requena-mesa_predicting_2018}. 
In this context, it is important to consider that ongoing and accelerated changes induced by global and climate change reshape the dynamics of the Earth system and respective data (making data through time non-stationary). 
Therefore, it becomes essential to explore solutions that streamline the adaptability and transferability of data-driven machine learning methods, ensuring their efficiency in extrapolating from existing and historical data to future circumstances.

%
%
Machine learning and computer vision are exerting a progressively increasing influence across various domains, including forest monitoring, as elaborated in Section \ref{sec:ml_challenges}. Strategies related to model generalization, learning schemes, and forestry-based metrics are valuable for delving further into the challenges presented by forest biology.

Enhancing the generalization capabilities of models involves better adaptation to diverse spatial and temporal domains, encompassing different forests, sensors, and resolutions. To achieve this, machine learning strategies will be explored, focusing on leveraging existing datasets through weakly-supervised (see Sec.~\ref{sec:ml_weakly}) or few-shot (see Sec.~\ref{sec:ml_fewshot} and \ref{sec:ml_zeroshot}) learning approaches. Moreover, hybrid-models, which integrate physical knowledge (see Sec.~\ref{sec:ml_metrics}), or space-for-time substitutions, which enable to learn temporal dynamics from spatial dynamics, may greatly advance our capabilities to design robust data-driven machine learning applications for monitoring and forecasting in a non-stationary world.
In line with this perspective, the \textbf{OpenForest} dynamic catalogue could serve as a suitable reference to consistently enhance and refine models using the latest data.

Implementing active learning methods (see Sec.~\ref{sec:ml_active}) can significantly optimize the process of generating annotations for future datasets. As datasets continue to grow in size, self-supervised learning methods (see Sec.~\ref{sec:ml_ssl}) offer a valuable perspective to learn meaningful representations in deep learning algorithms for forest monitoring without relying heavily on manual annotations.

Incorporating multi-modal and multi-task computer vision architectures into forest monitoring presents an intriguing opportunity to capitalize on task complementarity. 
For instance, by predicting multiple foliage traits from hyperspectral data, a model can learn the covariance among different traits and, hence, provide more robust estimates for challenging traits based on their relation with more accurately predicted ones \cite{cherif_spectra_2023, schiller_deep_2021}. 
An additional area of potential research could involve enhancing carbon stock estimation by simultaneously predicting both tree species and height.

Foundation models (see Sec.~\ref{sec:ml_foundation_models}) have demonstrated remarkable capabilities in managing various modalities, such as LiDAR, RADAR, and hyperspectral data, with varying spatial and temporal resolutions. These models remain task-agnostic and can achieve high zero-shot performances, making their pretraining a challenging yet promising endeavor for forest monitoring. Once pretrained, they can be adapted to multiple other tasks in this domain.
While relying on the complementarity of large scale multi-modal and multi-tasks datasets, research on foundation models for forest monitoring worldwide would benefit from the \textbf{OpenForest} catalogue dynamically enriched by the community.

Section \ref{sec:review} provides a comprehensive review of open-source forest datasets, categorized according to specific criteria and identified scales. These datasets are grouped in \textbf{OpenForest}, a dynamic catalogue open for updates from the community. The aim is to foster communication, inspire new applications of machine learning in forest monitoring, and motivate advancements in this field.

Datasets, as prerequisite of machine learning applications, commonly exhibit a lack of geographical representativeness, particularly noticeable in African and Asian regions, as depicted in Figure \ref{fig:distributions}. 
Whenever feasible, ideal datasets would preferably align multi-modal recordings based on their temporal and spatial resolutions while offering annotations in the highest available resolution.
As demonstrated in remote sensing \cite{lacoste_geo-bench_2023}, aggregating numerous diverse forest datasets could foster research in developing specialized foundation models for effective forest monitoring

The \textbf{OpenForest} catalogue, in addition to providing a list of open access datasets, will also curate information about data providers\footnote{See related information at \url{https://github.com/RolnickLab/OpenForest}.}, a crucial resource for generating well-structured datasets that cater to specific needs. Citizen-generated data, such as curated on OpenAerialMap\footnote{\url{https://openaerialmap.org/}} or GBIF\footnote{\url{https://www.gbif.org/}}, holds significant value since it integrates information from across the globe, making it ideal for self-supervised learning. The data provider list will also be frequently updated to integrate most recent initiatives. For instance, incorporating data from the Biomass mission\footnote{\url{https://www.esa.int/Applications/Observing_the_Earth/FutureEO/Biomass}} as soon as possible into future datasets is essential. The mission's provision of P-band SAR data will greatly benefit worldwide forest tomography \cite{berenger_deep_2023}, advancing our comprehension of forest carbon stock and its dynamics.
Exploring its potential can lead to the creation of valuable and structured datasets.

Recordings from aerial data, especially UAVs, gain momentum by offering promising prospects, with more affordable and easier-to-pilot vehicles equipped with higher-resolution sensors. Leveraging UAV technology allows for high-resolution forest analysis, even in remote or inaccessible areas, and can moreover advance large-scale assessments by its integration with Earth observation satellite missions \citep{schiefer_uav-based_2023}.

As more and more datasets are released at various scales, the \textbf{OpenForest} catalogue offers the opportunity to centralize these information with details. It will help to motivate research in bridging the gab between scales, sensors and resolutions while hopefully motivate collaborations between researchers.

\begin{Backmatter}

\paragraph{Acknowledgments} The authors are grateful for the valuable feedback of O.~Sonnentag.

\paragraph{Funding Statement}
This work was funded through the IVADO program on ``AI, Biodiversity and Climate Change'' and the Canada CIFAR AI Chairs program. 
It was also funded through the German Research Foundation (DFG) under the project PANOPS (project number 504978936) and BigPlantSens (project number 444524904).

\paragraph{Competing Interests}
None

\paragraph{Data Availability Statement}
The \textbf{OpenForest} catalogue is available and open to contributions in the following repository: \url{https://github.com/RolnickLab/OpenForest}.

\paragraph{Ethical Standards}
The research meets all ethical guidelines, including adherence to the legal requirements of the study country.

\paragraph{Author Contributions}
Conceptualization: A.O.; D.R. Methodology: A.O.; T.K.; E.L.; D.R. Data curation: A.O. Data visualisation: A.O. Writing original draft: A.O.; T.K.; E.L. Writing - Review \& Editing: A.O.; T.K.; E.L.; D.R.. All authors approved the final submitted draft.

\bibliographystyle{apalike}

\bibliography{openforest_cleaned}

\end{Backmatter}

\end{document}